\documentclass[lettersize,journal]{IEEEtran}

\usepackage[switch]{lineno}

\usepackage{amsmath,amsfonts,amsthm}
\usepackage{algorithmic}
\usepackage{algorithm}
\usepackage{array}
\usepackage[caption=false,font=normalsize,labelfont=sf,textfont=sf]{subfig}
\usepackage{textcomp}
\usepackage{stfloats}
\usepackage{url}
\usepackage{verbatim}
\usepackage{graphicx}
\usepackage{cite}
\usepackage{arydshln}

\usepackage{diagbox}

\usepackage{rotating}

\usepackage{xcolor}
\usepackage{enumitem}

\usepackage{xspace}
\newcommand{\nc}{\textit{NC}\xspace}
\newcommand{\nh}{\textit{NH}\xspace}

\theoremstyle{definition}
\newtheorem{definition}{\textbf{Definition}}
\newtheorem{proposition}{\textbf{Proposition}}
\newtheorem{observation}{\textbf{Observation}}

\begin{document}
% \linenumbers
\title{Clarify Confused Nodes via Separated Learning}

\author{
Jiajun Zhou, 
Shengbo Gong, 
Xuanze Chen, 
Chenxuan Xie, 
Shanqing Yu, 
\\
Qi Xuan, \IEEEmembership{Senior Member, IEEE}, 
Xiaoniu Yang
\thanks{This work was supported in part by the Key Research and Development Program of Zhejiang under Grants 2022C01018, in part by China Post-Doctoral Science Foundation under Grant 2024M762912, in part by the Post-Doctoral Science Preferential Funding of Zhejiang Province of China under Grant ZJ2024060, in part by the National Natural Science Foundation of China under Grant U21B2001, and in part by the Key Technology Research and Development Program of Hangzhou under Grant 2024SZD1A23. \emph{(Corresponding authors: Jiajun Zhou and Shanqing Yu.)}}
\thanks{Jiajun Zhou is with the Institute of Cyberspace Security, College of Computer Science and Technology, Zhejiang University of Technology, Hangzhou 310023, China, and also with the Binjiang Institute of Artificial Intelligence, ZJUT, Hangzhou 310056, China (e-mail: jjzhou@zjut.edu.cn).}
\thanks{Shengbo Gong, Xuanze Chen, Chenxuan Xie, Shanqing Yu and Qi Xuan are with the Institute of Cyberspace Security, Zhejiang University of Technology, Hangzhou 310023, China, and also with the Binjiang Institute of Artificial Intelligence, ZJUT, Hangzhou 310056, China (e-mail: yushanqing@zjut.edu.cn).}
\thanks{Xiaoniu Yang is with the National Key Laboratory of Electromagnetic Space Security, Jiaxing 314033, China (e-mail: yxn2117@126.com).}
\thanks{Co-first authors: Jiajun Zhou and Shengbo Gong.}
}

\markboth{Journal of \LaTeX\ Class Files,~Vol.~14, No.~8, August~2021}%
{Shell \MakeLowercase{\textit{et al.}}: A Sample Article Using IEEEtran.cls for IEEE Journals}


\maketitle

% \IEEEdisplaynontitleabstractindextext

% \IEEEpeerreviewmaketitle

\begin{abstract}
    Graph neural networks (GNNs) have achieved remarkable advances in graph-oriented tasks.
    However, real-world graphs invariably contain a certain proportion of heterophilous nodes, challenging the homophily assumption of traditional GNNs and hindering their performance.
    Most existing studies continue to design generic models with shared weights between heterophilous and homophilous nodes. Despite the incorporation of high-order messages or multi-channel architectures, these efforts often fall short. A minority of studies attempt to train different node groups separately but suffer from inappropriate separation metrics and low efficiency.
    In this paper, we first propose a new metric, termed Neighborhood Confusion (\textit{NC}), to facilitate a more reliable separation of nodes. We observe that node groups with different levels of \textit{NC} values exhibit certain differences in intra-group accuracy and visualized embeddings.
    These pave the way for \textbf{N}eighborhood \textbf{C}onfusion-guided \textbf{G}raph \textbf{C}onvolutional \textbf{N}etwork (\textbf{NCGCN}), in which nodes are grouped by their \textit{NC} values and accept intra-group weight sharing and message passing. 
    Extensive experiments on both homophilous and heterophilous benchmarks demonstrate that our framework can effectively separate nodes and yield significant performance improvement compared to the latest methods.
    The source code is available in \url{https://github.com/GISec-Team/NCGNN}.
\end{abstract}
    

\section{Introduction}
\IEEEPARstart{G}{raph} structured data can effectively model real-world interactive systems and has received considerable attention recently. Graph neural networks (GNNs) have been proven to be powerful in many graph-related applications, such as anomaly detection~\cite{ma2021comprehensive,luo2021future,ahmed2021graph}, fraud detection~\cite{cheng2020graph,zhou2022behavior,10490264} and recommendation systems~\cite{lyu2022knowledge,melton2023muxgnn,cai2021line,yan2024federated}. Traditional GNNs~\cite{kipf2016semi,hamilton2017inductive,velivckovic2017gat} are adept at learning node representations in graphs where the vast majority of nodes are homophilous, i.e., \textit{nodes are primarily connected to nodes with same labels.} However, real-world graphs often contain heterophilous nodes~\cite{liu2022ud} and are known as heterophilous graphs when the proportion of these nodes is high, which challenges the homophily assumption of traditional GNNs and resulting in diminished performance~\cite{zheng2022survey,zheng2023node,wang2021graph}. In heterophilous graphs, the patterns of connections between nodes are usually complex and varied, rather than being limited to connections with similarly labeled nodes. This is referred to as the \textbf{heterophily problem}, and recent studies~\cite{mao2023demystifying,li2023restructuring} have highlighted that the key to addressing this problem is to focus on the diversity of nodes in the graph. Inspired by this, it's natural to designing flexible model architectures for characterizing diverse nodes, whether homophilous or heterophilous.

Most recent studies endeavor to design general models that still enable weight sharing between homophilous and heterophilous nodes. However, even with the incorporation of higher-order neighbors~\cite{chien2020adaptive,he2021bernnet,chen2020gcnii,zhou2024pathmlp} or multi-channel architectures~\cite{luan2022acm,bo2021fagcn}, these models fail to perform consistently well across all evaluation datasets. 
This may be due to two reasons: 
1) \emph{these methods do not fundamentally alter the message-passing mechanism that essentially conforms to homophily assumption, thereby introducing noise}~\cite{wang2022hog}; 
2) \emph{nodes within the same graph may exhibit significant behavioral differences due to distribution bias}~\cite{wuenergy,li2022graphde}. 
For example, categorizing an interdisciplinary paper relies on more specific criteria compared to a paper confined to a single discipline~\cite{bornmann2008citation}.
\emph{Therefore, sharing weights universally across all nodes in a real-world graph is impractical, prompting the necessity for a metric to \textbf{separate} nodes with different characteristics.}
While numerous studies have explored the separation of neighbors~\cite{du2022gbk,yan2022two} or spectral signals~\cite{luan2022acm,bo2021fagcn}, the separation of target nodes or transformation weights, which we refer to as \textbf{separated learning}, remains under-explored. 
Only a minority of studies~\cite{liu2022ud,du2022gbk} have focused on this separating scheme, but \emph{they suffer from incorrect separation metrics and low efficiency.}
\begin{figure}[!htb]
    \centering
    \includegraphics[width=0.9\linewidth]{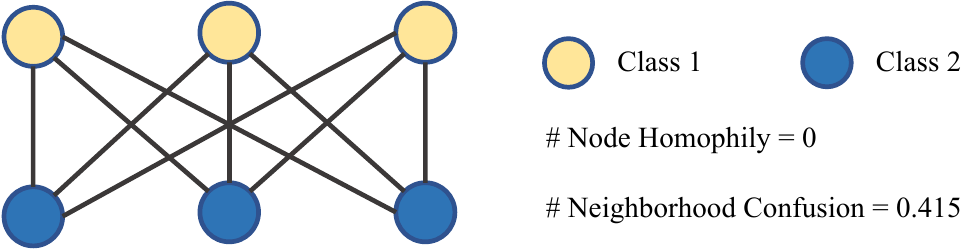}
    \caption{Illustration of the paradox arising from the node homophily metric.}
    \label{fig:case1}
\end{figure}

Beginning with the classical Node Homophily (\textit{NH}) metric~\cite{pei2019geom} derived directly from the definition of homophily or heterophily, we analyze the bipartite graph scenario as depicted in Fig.~\ref{fig:case1}. According to the node homophily metric defined in Eq.~(\ref{eq: NH}), the homophily of any node in the bipartite graph is zero. Nevertheless, graph convolutional network (GCN) can still perform well in such extremely heterophilous scenario, presenting a paradox~\cite{ma2021necessity}.
This phenomenon has been widely observed in recent studies~\cite{ma2021necessity,zhu2020h2gcn,luan2022acm}, \textit{yet they neither ease this anomaly directly with their models nor explore it on a wider range of real-world graphs.} 
Moreover, analytical papers~\cite{zhang2022degradation,oono2019graph,li2018deeper} reach the consensus that message passing with too many layers leads to over-smoothing or indistinguishability on node representations, highlighting the importance of distinguishability over consistency with central nodes for model performance.
Therefore, we propose a new metric termed Neighborhood Confusion (\textit{NC}) and detail its superiority over \textit{NH}. Our proposed framework is guided by this metric for node separation, further validating the effectiveness of \textit{NC}.
\textbf{Our contributions can be summarized as follows:}
\begin{itemize}[leftmargin=15pt]
    \item We propose a new metric \textit{NC} based on observations of results on GCN from multiple perspectives. And we prove that it is an appropriate metric for node separation and can be computed in real time during model training due to its low computational complexity.
    \item We design a novel framework named \textbf{N}eighborhood \textbf{C}onfusion-guided \textbf{G}raph \textbf{C}onvolutional \textbf{N}etwork (\textbf{NCGCN}), which is the first to break the weight sharing assumption of GCN. In our framework, the metric estimation and model inference are jointly optimized in an end-to-end manner. 
    Furthermore, we demonstrate the potential of our framework to serve as a general enhancement plug-in by substituting its backbone with an alternative message passing neural network.
    \item Our framework achieves SOTA in ten widely used benchmarks, surpassing the latest methods tailor-made for these datasets while maintaining the simplicity of GCN. Extensive experiments demonstrate that our proposed metric \textit{NC} plays a key role during model training.
\end{itemize}

\section{Related Work}
\subsection{Heterophily Problem}
To address the heterophily problem, several studies resort to the spectral theory. 
GPRGNN~\cite{chien2020adaptive} adaptively optimizes the Generalized PageRank weights to control the contribution of propagation at each layer.
Mixhop~\cite{abu2019mixhop}, H2GCN~\cite{zhu2020h2gcn} and Snowball~\cite{luan2019snowball} combine ego- and neighbor-embedding or multi-order intermediate representations, while FAGCN~\cite{bo2021fagcn} and ACM-GCN~\cite{luan2022acm} utilize multi-channel local filters to capture local information. 
Limited by the globality of spectral theory, these methods are inherently unable to separate nodes and filter neighbors. Additionally, the correlation between high frequency and heterophily lacks reliability.

Spatial methods allow for fine-grained adaptations of message passing and transformation rules~\cite{zhang2019survey}.
Geom-GCN~\cite{pei2019geom} and WRGAT~\cite{suresh2021wrgat} rewire graphs to aggregate neighbors with structural similarities in latent space. SNGNN~\cite{zou2023similarity} and GGCN~\cite{yan2022two} construct a similarity matrix, drawing upon embeddings from the preceding layer, to separate the dissimilar neighbors. These methods require computing new topologies or similarities and thus are time and memory consuming. 

Several studies also introduce new metrics, e.g., \textit{aggregated similarity score}~\cite{luan2022acm}, \textit{local assortativity}~\cite{suresh2021wrgat}, 2NCS~\cite{cavallo2023NCS}, density-aware homophily~\cite{li2023restructuring} and \textit{neighbor homophily}~\cite{cagnn}. However, these are often not effectively leveraged to guide model training, owing to high computational complexity or incompatibility with message passing layers. 
It reduces both the rationality of proposed metrics and the theoretical support of proposed models. 
To the best of our knowledge, we are the first to incorporate the proposed metric into model training in an end-to-end manner.

\subsection{Related Work on Label Reuse}
Label distribution is the direct reason why the graphs are categorized as homophilous or heterophilous, so the label reuse~\cite{wang2021bag} technique, i.e., utilizing information from labels, pseudo labels or soft assignment prediction, can be helpful. 
GBKSage~\cite{du2022gbk} groups neighbors based on whether they belong to the same class as the target nodes, soft-separating neighbors through a gating mechanism. 
GHRN~\cite{xiaoliu} calculates similarity of post-aggregation pseudo labels and then filters connections. 
CPGNN~\cite{zhu2021cpgnn} and HOG~\cite{wang2022hog} propagate pseudo labels to achieve an affinity matrix for calibrating the training process. Both of them use label propagation that still conforms to the homophily assumption. 
LWGCN~\cite{lwgcn} separates each class of neighbors and concatenates them. This approach can introduce considerable nulls, especially when a node has fewer neighbors but the graph has many classes. 
Furthermore, the majority of the aforementioned methods compromise the simplicity inherent in label information, introducing complexity by computing feature similarities. In contrast, our \textit{NC} metric demonstrates that label distribution alone is sufficient to effectively guide models without being overly simplistic or sensitive.

\subsection{Related Work on Separated Learning}
Separated learning involves applying distinct learning rules to target nodes.
UDGNN~\cite{liu2022ud} uses continual learning to achieve this separation. However, the metric \textit{NH} they use is inappropriate, as we mentioned above.
One possible reason for the gap between low \textit{NC} and high \textit{NC} groups is that they are sampled from different distributions. 
GraphDE~\cite{li2022graphde} tries to identify in-distribution instances and treats synthesized out-of-distribution instances as noise, stopping gradient descent for the latter. 
In contrast, assuming that out-of-distribution instances belong to the other distribution, our approach learns the separated part with the other learner.

\section{Preliminaries}
\subsection{Notations} 
An attributed graph can be represented as $G=(V,E, \boldsymbol{X}, \boldsymbol{Y})$, where $V$ and $E$ are the sets of nodes and edges respectively, $\boldsymbol{X}\in \mathbb{R}^{|V|\times d}$ is the node feature matrix, and $\boldsymbol{Y}\in \mathbb{R}^{|V|\times C}$ is the node label matrix. 
Here we use $|V|$, $d$, $C$ to denote the number of nodes, the dimension of the node features, and the number of node classes, respectively.
Each row of $\boldsymbol{X}$ (i.e., $\boldsymbol{x}_i$) represents the feature vector of node $v_i$, and each row of $\boldsymbol{Y}$ (i.e., $\boldsymbol{y}_i$) represents the one-hot label of node $v_i$.
The structure elements $(V, E)$ can also be represented as an adjacency matrix $\boldsymbol{A}\in\mathbb{R}^{|V|\times|V|}$ that encodes pairwise connections between the nodes, whose entry $\boldsymbol{A}_{ij} = 1$ if there exists an edge between $v_i$ and $v_j$, and $\boldsymbol{A}_{ij} = 0$ otherwise.
Based on the adjacency matrix, we can define the degree distribution of $G$ as a diagonal degree matrix $\boldsymbol{D} \in \mathbb{R}^{|V|\times |V|}$ with entries $\boldsymbol{D}_{ii} = \sum_{j=1}^{|V|} \boldsymbol{A}_{ij}$ representing the degree value of $v_i$. 
We denote the set of neighbors of node $v_i$ within $k$ hops as $\mathcal{N}_i^k=\{v_j \mid  1\leq  \operatorname{ShortestPath}(v_i,v_j)\leq k \}$.

\subsection{Node Homophily}\label{sec:NH}
The most widely used metric is Node Homophily (\textit{NH})~\cite{pei2019geom}. For a target node $v_i$, its \textit{NH} value is defined as: 
\begin{equation}\label{eq: NH}
    N H_i=\frac{\left|\left\{v_j \mid v_j \in \mathcal{N}_i^1, y_j=y_i\right\}\right|}{\left|\mathcal{N}_i^1\right|}
\end{equation}
where $y$ represents the node label. \textit{NH} represents the proportion of the direct neighbors that have the same label as the target node in the neighborhood. According to whether the \textit{NH} is greater or less than 0.5, nodes are categorized as homophilous and heterophilous~\cite{liu2022ud}. The average \textit{NH} computed from the nodes across the entire graph is usually regarded as a metric for assessing graph homophily.

\subsection{Graph Convolutional Network} 
The standard GCN with two layers proposed in \cite{kipf2016semi} can be formulated as: 
\begin{equation}
    \boldsymbol{B} = \operatorname{softmax}(\operatorname{norm}(\boldsymbol{A}) \cdot \operatorname{ReLU}(\operatorname{norm}(\boldsymbol{A})\cdot \boldsymbol{X} \boldsymbol{W}_1)\cdot \boldsymbol{W}_2)
\end{equation}
where $\boldsymbol{B}$ is the soft assignment of label predictions, and $\operatorname{norm}(\cdot)$ is the symmetric normalization operation with 
$\operatorname{norm}(\boldsymbol{A})=(\boldsymbol{D}+\mathbf{I})^{-1/2}(\boldsymbol{A}+\boldsymbol{I})(\boldsymbol{D}+\boldsymbol{I})^{-1/2} $.
It can be regarded as two layers of \textit{message passing}, represented by $\operatorname{norm}(\boldsymbol{A})\cdot\boldsymbol{X}$, followed by \textit{transformation} $\boldsymbol{W}_1/\boldsymbol{W}_2$. Note that the message passing is indiscriminate to all neighbors and the transformation shares weight among all nodes.

%%%%%%%%%%%%%% Observation %%%%%%%%%
\section{Metrics and Observations}\label{sec:NC}
Heterophily is harmful to traditional GNNs, as different classes of nodes will be inappropriately mixed during message aggregation, leading to the indistinguishability of node representations~\cite{zhu2020h2gcn}.
However, GCN still performs well on bipartite graphs, which are completely heterophilous under the definition of \textit{NH}, bringing paradox~\cite{ma2021necessity}.
As shown in Fig.~\ref{fig:case1}, $\textit{NH}=0$ means that \textit{NH} metric considers the nodes in the bipartite graph to be extremely heterophilous.
To alleviate this paradox, some new metrics have been proposed, such as \textit{aggregated similarity score}~\cite{luan2022acm}, which is calculated by the similarity of intermediate representations, but with high computational complexity.

\subsection{Neighborhood Confusion} 
Rethinking the aforementioned issue, we contend that the perceived homophily or heterophily is merely a superficial phenomenon. The crux influencing model performance lies in the difficulty of distinguishing nodes, intrinsically tied to the label distribution surrounding target nodes, as well as the non-independent nature of message passing.
Therefore, we propose a new metric named Neighborhood Confusion (\textit{NC}), which can measure the label diversity in the node neighborhood.
\begin{definition}{\emph{\textbf{Neighborhood Confusion.}}}
    For a target node $v_i$, its Neighborhood Confusion is defined as:
    \begin{equation}\label{eq:NC}
        \begin{array}{c}
            NC_i^{(k)} =-\frac{1}{\log |\mathcal{C}|} \cdot \log \frac{\left| \left\{v_j \ \mid\  v_j \in \mathcal{N}_i^k \cup \left\{v_i\right\}, \  y_j = c_{\ast}\right\} \right|}{\left| \mathcal{N}_i^k \cup \left\{v_i\right\} \right|} \vspace{0.5em}\\
            \text{with} \quad c_{\ast} = \underset{c\in \mathcal{C}}{\arg\max}\left| \left\{v_j \mid v_j \in \mathcal{N}_i^k \cup \left\{v_i\right\}, y_j = c \right\}\right| \\
        \end{array}
    \end{equation}
where $\mathcal{C}=\{1,2,\dots,C\}$ is the set of label classes, $\frac{1}{\log |\mathcal{C}|}$ normalizes \textit{NC} to $[0,1]$, and higher value indicates a greater degree of label confusion within target neighborhood. 
The receptive field of \textit{NC} can be 1- or 2-hop by setting $k=1$ or $2$, which is adaptively configured across different datasets.
\end{definition}

\begin{proposition}
    \emph{\textit{NC} metric can outline the joint label distribution in the $k$-hop ego-net of node $v_i$, and can be approximated as a loose form of joint label distribution entropy in highly heterophilous scenarios, but with reduced computational complexity by focusing on the most frequent label in the ego-net.}
\end{proposition}

\begin{proof}
    The joint probability that a target node $v_i$ and its neighbors in the $k$-hop neighborhood $\mathcal{N}_i^k$ have label $c$ can be denoted as $p(y_i=c, Y_{\mathcal{N}_i^k}=c)$, where $Y_{\mathcal{N}_i^k}$ is subject to the label distribution of neighbors. We will abbreviate the joint probability as $p(c)$ for simplicity. Then the joint entropy of the label distribution around (and include) target node $v_i$ is:
    \begin{equation}\label{eq:Hx}
        \begin{aligned}
        &H(y_i, Y_{\mathcal{N}_i^k})=-\sum_{c=1}^{|\mathcal{C}|} p(c)\cdot\log p(c)\\
       &\text{with} \quad p(c)= \frac{|\{v_j \mid v_j \in \mathcal{N}_i^k \cup \{v_i\},\  y_j = c \}|}{|\mathcal{N}_i^k \cup \{v_i\}|}
        \end{aligned}
    \end{equation}
    Since \nc is based on the most frequent label, we express it as a loose form of joint entropy.
    By defining $q(c) = p(c)$ and $p_{\max}^\ast = \max_{c\in \mathcal{C}} p(c)$, we have: 
    \begin{equation}\label{eq:proof1}
        \begin{array}{l}
            \text{\emph{NC}}_i^{(k)}=-\frac{1}{\log |\mathcal{C}|} \cdot \log \frac{\left| \left\{v_j \mid v_j \in \mathcal{N}_i^k \cup \{v_i\},\  y_j = c_{\ast}\right\} \right|}{\left| \mathcal{N}_i^k \cup \{v_i\} \right|}\vspace{0.5em}\\
            = -1 \cdot \frac{1}{\log |\mathcal{C}|}\cdot \log p_{\max}^\ast\vspace{0.5em}\\
            = -\left(\sum_{c=1}^{|\mathcal{C}|} {q(c)}\right) \cdot \frac{1}{\log |\mathcal{C}|}\cdot \log p_{\max}^\ast\vspace{0.5em}\\
            = - \frac{1}{\log |\mathcal{C}|}\left( q(1)\cdot \log p_{\max}^\ast + \cdots + q({|\mathcal{C}|})\cdot \log p_{\max}^\ast \right) \vspace{0.5em}\\
            \leq - \frac{1}{\log |\mathcal{C}|}\left( q(1)\cdot \log p(1) + \cdots + q({|\mathcal{C}|})\cdot \log p({|\mathcal{C}|}) \right) \vspace{0.5em}\\
            = - \frac{1}{\log |\mathcal{C}|} \sum_{c=1}^{|\mathcal{C}|} q(c)\cdot\log p(c)\vspace{0.5em}\\ 
            =\frac{1}{\log |\mathcal{C}|} \cdot H(y_i, Y_{\mathcal{N}_i^k})
        \end{array}
    \end{equation}
\noindent Thus, as the label distribution in the ego-net becomes more uniform and diverse (i.e., the environment becomes more heterophilous), \nc can be interpreted as an approximation of joint entropy $H(y_i, Y_{\mathcal{N}_i^k})$. The range of $H(y_i, Y_{\mathcal{N}_i^k})$ is determined to be $\left[0,\log |\mathcal{C}|\right]$ based on the Maximum Entropy Principle. Consequently, \nc can be normalized within the interval $[0,1]$ by dividing $\log |\mathcal{C}|$.
While joint entropy uses the full label distribution, \textit{NC} only considers the most frequent label, reducing the complexity of the computation.
\end{proof}

\begin{figure}[!htb]
    \centering
    \includegraphics[width=0.7\linewidth]{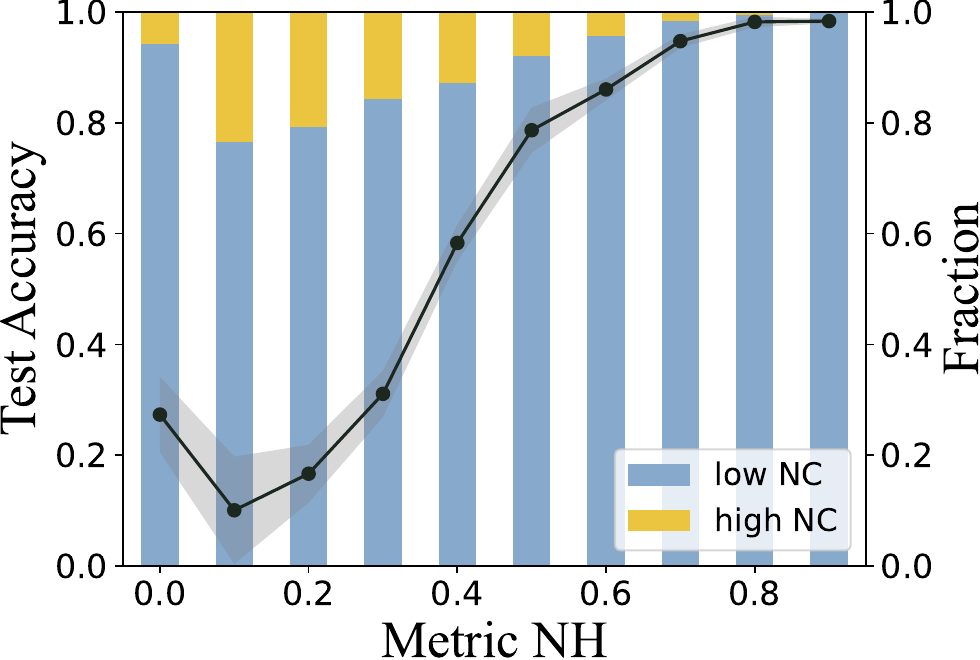}
    \caption{
    Observational experiment on the \texttt{Coauthor CS} dataset. Nodes are divided into 10 groups based on the \nh metric, and the test accuracy of GCN across different node groups varies with \nh value and the proportion of high-\nc nodes. In the group with the lowest \nh value, the abnormally high accuracy benefits from the high proportion of low-\nc nodes.
    }
    \label{fig:fraction}
\end{figure}

\begin{figure*}[!htb]
    \centering
    \includegraphics[width=0.85\linewidth]{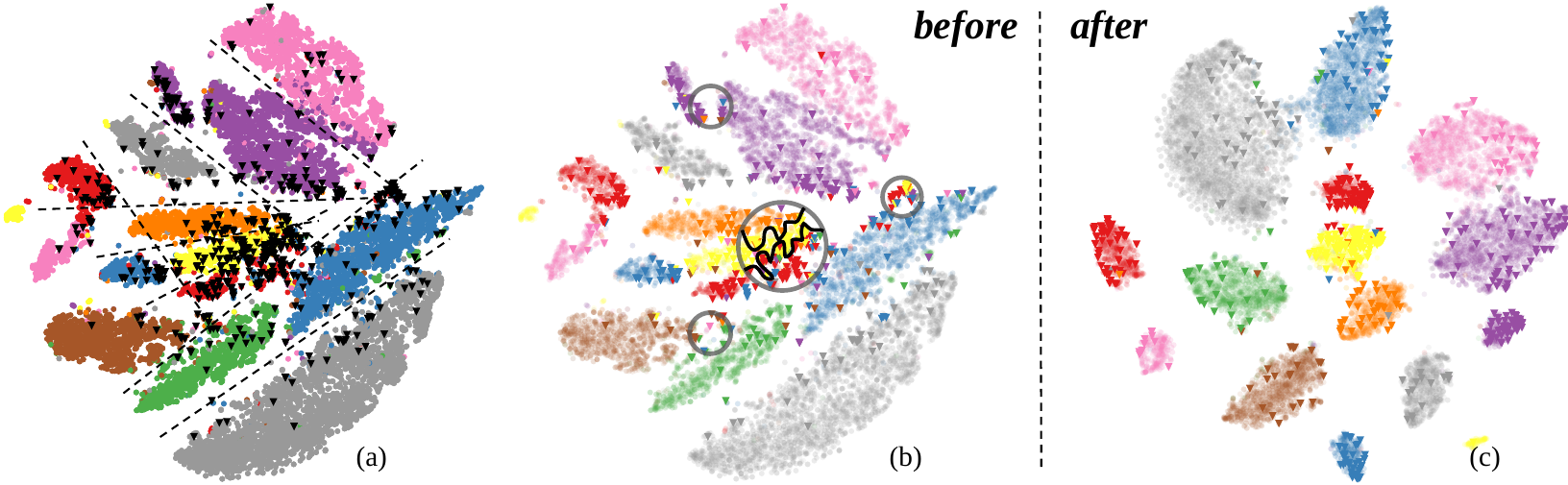}
    \caption{Visualization of embeddings learned by GCN on \texttt{Coauthor CS} dataset before and after guidance by \nc metric. Triangles and dots represent high-\nc and low-\nc nodes, respectively.
    Observations: 
    (a) The black triangles representing high-\nc nodes are primarily distributed at the edges of clusters and tend to gather at the intersection of the decision boundaries, as indicated by the black dashed lines;
    (b) The circled clusters, composed of different categories of high-\nc nodes, require more complex decision boundaries to be effectively separated.
    (c) After optimization guided by the \nc metric, high-\nc nodes are distanced from the decision boundaries and are more evenly distributed within the clusters of their corresponding classes.
    }
    \label{fig:scatter}
\end{figure*}

\subsection{Observations}
Rethinking the scenario of bipartite graph, as depicted in Fig.~\ref{fig:case1}, $NC^{(1)}=0.415$ suggests that our metric perceives the label distribution of node neighborhoods to have a low level of confusion, which is quite different from the result of \nh.
Then, we will further explain why \nc surpasses \nh in separating nodes from two observations. As an illustrative analysis, we train a standard GCN (2 layers with 64 hidden units, 0.5 dropout rate, and ReLU) on the public benchmark \texttt{Coauthor CS}~\cite{shchur2018pitfalls}, with the 2-hop \nc and a threshold of 0.4 to separate high-\nc nodes from low-\nc nodes. 

\begin{observation}
    We first calculate the \nh values for nodes in the validation set and divide all nodes into ten groups according to the deciles of their \nh values. Fig.~\ref{fig:fraction} reports the average test accuracy of each group on 10 runs, along with the proportion of high-\nc and low-\nc nodes in each group. Consistent with previous studies, GCN achieves exceptionally high test accuracy within the group with the lowest \nh, indicating that nodes in this group are not the least distinguishable. In conjunction with our \nc metrics, this phenomenon can be easily explained by the negative correlation between test accuracy and the proportion of high-\nc nodes. Specifically, the proportion of high-\nc nodes in the group with \nh of 0.0 is lower than that in the group with \nh of 0.1, thereby resulting in a higher test accuracy. When the \nh is 0.1 or higher, as it increases, the test accuracy rises as the proportion of high-\nc nodes decreases. Such observation suggests that the \nc metric can characterize the distinguishability of nodes to a certain extent. 
    The results on more benchmark datasets are presented in Fig.~\ref{fig:nh-acc}.
\end{observation}

\begin{proposition}
    \emph{\nc metric not only measures the distinguishability of nodes in a graph but also provides insight into the classification uncertainty for each node. Nodes with higher \nc values are generally more challenging to classify due to the heterophily of their neighborhoods. The classification error rate for such nodes can be theoretically bounded using \nc.}
\end{proposition}

\begin{proof}
    To understand how neighborhood confusion impacts node classification, we first turn to conditional entropy $H(y_i | Y_{\mathcal{N}_i^k})$, which represents the expected uncertainty in the target node's label $y_i$ given the labels of its neighbors $Y_{\mathcal{N}_i^k}$. In other works, $H(y_i | Y_{\mathcal{N}_i^k})$ reflects how much confusion remains about the target node's label even after knowing its neighborhood's labels. The conditional entropy $H(y_i | Y_{\mathcal{N}_i^k})$ can be formulated as follows:
    \begin{equation}
        H(y_i| Y_{\mathcal{N}_i^k})= -\sum_{c=1}^{|\mathcal{C}|} p(y_i=c| Y_{\mathcal{N}_i^k})\cdot\log p(y_i=c| Y_{\mathcal{N}_i^k})
    \end{equation}
    where $p(y_i=c\mid Y_{\mathcal{N}_i^k})$ represents the conditional probability that node $v_i$ has label $c$ given its neighborhood labels $Y_{\mathcal{N}_i^k}$.
    
    Assuming that the label probability of the target node $v_i$ is determined by its neighborhood label distribution, we have:
    \begin{equation}
        p(y_i=c\mid Y_{\mathcal{N}_i^k}) = \frac{|\{v_j \mid v_j \in \mathcal{N}_i^k ,\  y_j = c \}|}{|\mathcal{N}_i^k|}
    \end{equation}
    For the most frequent label $c_\ast$ in the neighborhood, we have: 
    \begin{equation}
        \begin{array}{c}
           p(y_i=c_\ast\mid Y_{\mathcal{N}_i^k}) = \frac{|\{v_j \mid v_j \in \mathcal{N}_i^k ,\  y_j = c_\ast \}|}{|\mathcal{N}_i^k|} \vspace{0.5em}\\
           \text{with} \quad c_{\ast} = \underset{c\in \mathcal{C}}{\arg\max}\left| \left\{v_j \mid v_j \in \mathcal{N}_i^k, y_j = c \right\}\right| 
        \end{array}
    \end{equation}
    Let $p_{\max}=p(y_i=c_\ast\mid Y_{\mathcal{N}_i^k})$. The remaining $|\mathcal{C}|-1$ labels ($c\neq c_\ast$) occupy the proportion $1-p_{\max}$, and for simplicity, we assume they are uniformly distributed, such that each of these labels has a conditional probability as follows:
    \begin{equation}
        p(y_i=c\mid Y_{\mathcal{N}_i^k}) = \frac{1-p_{\max }}{|\mathcal{C}|-1}, \quad \forall c\neq c_\ast
    \end{equation}
    Substituting the above conditional probabilities into the definition of conditional entropy, we obtain:
    \begin{equation}\label{eq: tiaojian-pmax}
        H(y_i| Y_{\mathcal{N}_i^k})=-p_{\max } \log p_{\max }-\left(1-p_{\max }\right) \log \frac{1-p_{\max }}{|\mathcal{C}|-1}
    \end{equation}
    When the neighborhood size is sufficiently large, we can approximate $p_{\max}\approx p_{\max}^\ast$. This allows us to establish the connection between conditional entropy $H(y_i| Y_{\mathcal{N}_i^k})$ and \nc.

    According to the definition of \nc, we have:
    \begin{equation}
        \textit{NC}=-1 \cdot \frac{1}{\log |\mathcal{C}|}\cdot \log p_{\max}^\ast \approx -1 \cdot \frac{1}{\log |\mathcal{C}|}\cdot \log p_{\max}
    \end{equation}
    Then the conditional probability of the most frequent label is:
    \begin{equation}\label{eq: pmax nc}
        p_{\max} = 2^{-\textit{NC} \cdot \log |\mathcal{C}|}
    \end{equation}
    Here we consider the base of the logarithm to be 2. Then, substiituting Eq.~(\ref{eq: pmax nc}) into Eq.~(\ref{eq: tiaojian-pmax}) yields:
    \begin{equation}
        \begin{aligned}
            H(y_i| Y_{\mathcal{N}_i^k})&= \textit{NC} \cdot \log |\mathcal{C}| \cdot 2^{-\textit{NC} \cdot \log |\mathcal{C}|} \\
            &- (1-2^{-\textit{NC} \cdot \log |\mathcal{C}|})\cdot \log \frac{1-2^{-\textit{NC} \cdot \log |\mathcal{C}|}}{|\mathcal{C}|-1}
        \end{aligned}
    \end{equation}
    Viewing conditional entropy $H(y_i| Y_{\mathcal{N}_i^k})$ as a function of \nc, it is easy to prove that conditional entropy is positively correlated with \nc in the interval $\textit{NC}\in[0,1]$ (refer to Appendix~\ref{app:proof} for more details). In other words, as the neighborhood becomes more confused (with a higher \nc value), the conditional entropy increases, indicating greater uncertainty in predicting the label of the target node based on neighborhood information. Therefore, the \nc metric can measure the classification difficulty (or distinguishability) of nodes in a graph.

    Next, we apply Fano's Inequality~\cite{fano1961transmission} to establish a theoretical lower bound on the classification error rate $P_e$ based on the conditional entropy $H(y_i\mid Y_{\mathcal{N}_i^k})$ as follows: 
    \begin{equation}
        P_e \geq \frac{H(y_i\mid Y_{\mathcal{N}_i^k})-1}{\log |\mathcal{C}|} = \frac{f(\textit{NC})-1}{\log |\mathcal{C}|} \ \text{(Lower Bound)}
    \end{equation}
    We can see that the lower bound function on the right side remains positively correlated with \nc, i.e., an increase in \nc will lead to a larger lower bound on the classification error rate, indicating that nodes with higher \nc values experience greater classification uncertainty and pose challenges for accurate classification. Furthermore, the classification error rate can be theoretically bounded using \nc.
    The curve of conditional entropy and classification error bound versus \nc under different numbers of label categories is shown in Appendix~\ref{app:proof}.
\end{proof}

\begin{observation}
    Furthermore, we visualize the embedding obtained from the output of the first layer of the trained GCN in Fig.~\ref{fig:scatter}(a). The clusters corresponding to different classes can be roughly separated by decision boundaries depicted by black dashed lines. These black triangles symbolize high-\nc nodes, predominantly distributed proximate to the cluster boundaries.
    When different decision boundaries (black dashed lines) intersect, clusters comprising nodes with high-\nc values emerge. 
    The clusters circled in Fig.~\ref{fig:scatter}(b) encompass diverse categories of high-\nc nodes, thereby necessitating more complex decision boundaries for effective segregation.
    Test results show that the overall test accuracy is 93.99\%, with 95.91\% for the low-\nc group and only 57.86\% for the high-\nc group.
\end{observation}

These phenomena support the argument that high-\nc nodes confuse GCN. We describe this as nodes becoming \textbf{confused} when their surrounding neighbors hold diverse views. In such cases, nodes require clarification, that is, they are highly motivated to reconsider their raw features, discern messages from their noisy neighbors, and ultimately adopt targeted classification rules. Given the significant gap between high- and low-\nc nodes, it is natural to optimize them separately to fit different learning processes. With our NCGCN model for separating the two groups of nodes during learning, high-\nc nodes are distanced from the decision boundaries and are more evenly distributed within the clusters of their respective classes, as shown in Fig.~\ref{fig:scatter}(c). The test results show that the overall test accuracy improves to 95.67\%, with 95.90\% for the low-\nc group and 82.47\% for the high-\nc group, yielding a significant boost in the latter.

\begin{proposition}
    \emph{Node Confusion (NC) metric exhibits greater robustness compared to the Node Homophily (NH) metric. Unlike \nh, which focuses more on the label consistency between the target node and its neighborhood, \nc primarily considers the number of high-frequency labels. Assuming a sufficiently large number of neighbors, \nc tends to be more stable when label perturbations occur in the neighborhood, whereas \nh is more susceptible to uncontrollable effect, particularly when the label of the target node is disturbed.
    Moreover, compared to similarity-based metrics, \nc can be computed efficiently during model training. Compared to label-based metrics, it is free from the constraint of consistency with the target nodes, thereby allowing a larger range of neighbors to be counted.
    Compared to difficulty metrics, \nc does not require temporal tracking across multiple epochs, allowing for immediate node differentiation within a single training step. Compared to predictability metrics, \nc encompasses all nodes, including those with high complexity, rather than excluding low-predictability or out-of-distribution samples, enabling a more comprehensive approach to node classification.}
\end{proposition}

\begin{figure*}[htbp]
    \centering 
    \includegraphics[width=\linewidth]{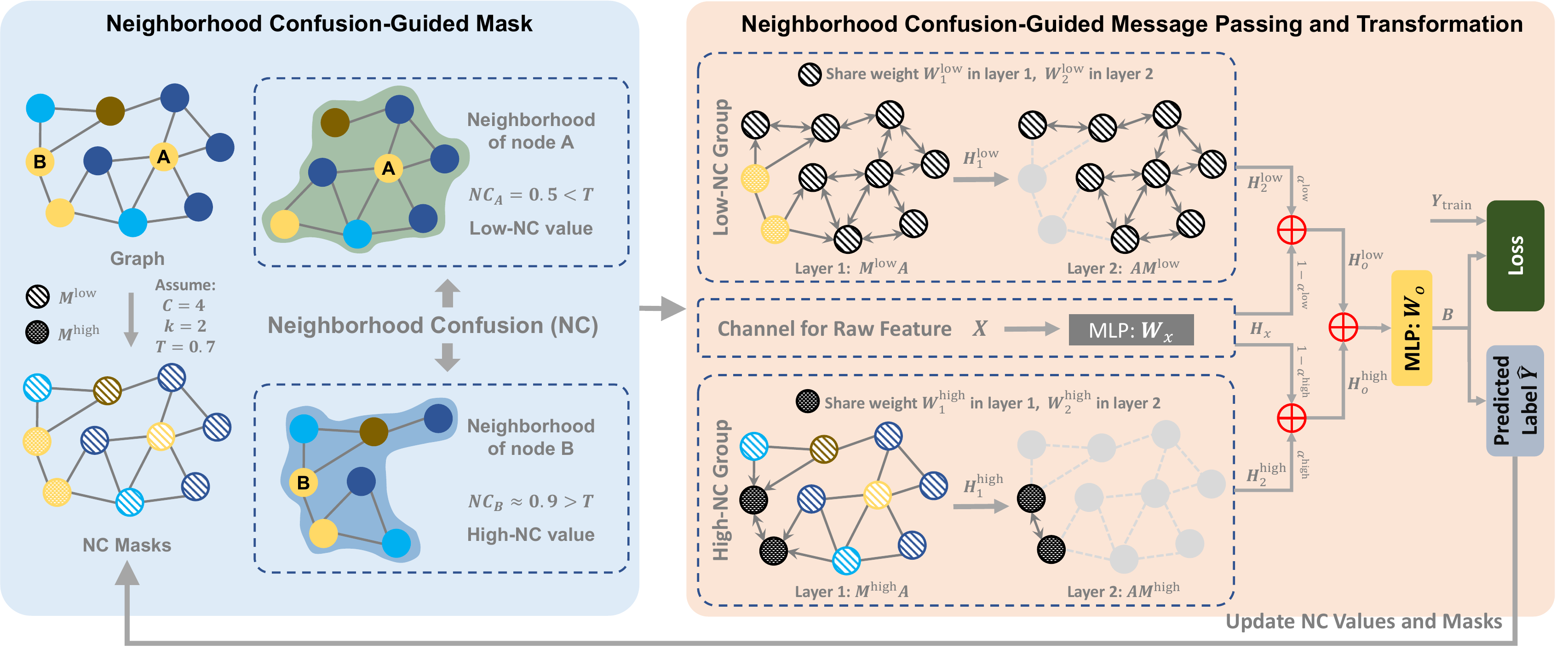}
    \caption{Illustration of NCGCN framework. The complete workflow proceeds as follows: 
    1) calculate \textit{NC} values of nodes and generate \textit{NC} masks, which can divide all nodes into high-\textit{NC} and low-\textit{NC} groups; 
    2) the two groups of nodes are trained with specific transformation weights and dropout rates in parallel. Message passing is constrained intra-group at the second layer. The raw features are adaptively added into the outputs. 
    3) the pseudo labels are used to update the \textit{NC} values of all nodes, and then the above two processes are repeated until the model converges.}
    \label{fig:framework}
\end{figure*}

%%%%%%%%%%%%%%%%%%%%%%%%%%%%%%%%%%%%%%%%%%%%%%%%%%%%%%%%%%%%%%%%%%%%%%
\section{Methodology}\label{sec:NCGCN}
Based on the observations, we can conclude that \nc metric can measure the distinguishability of nodes, thus being more suitable for node separation. 
Due to the huge gap between these two groups of nodes, it is presumed that they necessitate distinct learning processes tailored to their respective levels of distinguishability. In this section, we propose the Neighborhood Confusion-guided Graph Convolutional Network (NCGCN). Fig.~\ref{fig:framework} shows the overall framework.

\subsection{Neighborhood Confusion-guided Mask}\label{sec: NC-mask}
In our framework, we separate the nodes into two groups, i.e., high- and low-\nc nodes. Similar to ~\cite{xiaoliu,klicpera2019diffusion,zou2023similarity}, such absolute separation brings high computational efficiency. According to Eq.~(\ref{eq:NC}), the calculation of \nc metrics relies on label information. However, it is not permissible to use all the ground-truth labels during training, so \nc is calculated using pseudo labels, i.e., the model's predicted labels for the nodes are used in place of the ground-truth labels for metric computation.
Specifically, we initialize \nc values of all nodes as 0, and then update \nc values each time the validation accuracy reaches a new high. A threshold $T$ is used to separate nodes into high- and low-\nc groups, thus yielding the Neighborhood Confusion-guided Masks (\nc mask) by
\begin{equation}\label{eq: mask}
    \boldsymbol{M}^\text{low}_{ii}=\mathbb{I}\left(\textit{NC}_i \leq T\right), \quad
    \boldsymbol{M}^\text{high}_{ii}=\mathbb{I}\left(\textit{NC}_i > T\right)
\end{equation}
where $ \mathbb{I}(\cdot) $ is the indicator function. Both $\boldsymbol{M}^\text{low}$ and $\boldsymbol{M}^\text{high}$ are diagonal matrices.
Nodes with \nc values below (or above) $T$ are separated into low-\nc (or high-\nc) groups.

\begin{algorithm}[!htb]
    \caption{Training NCGCN}
    \label{alg:NCgcn}
    \begin{algorithmic}[1]
        \REQUIRE {Graph $G=(V,E,\boldsymbol{X},\boldsymbol{Y})$, maximum epochs $E$, early stopping patience $S$.}
        \ENSURE {Test accuracy $Acc_\text{test}$.}
            \STATE Initialize best validation accuracy: $Acc_\text{max} \gets 0$;
            \STATE Initialize \textit{NC} value for $\forall v_i \in V $: $\textit{NC}_i \gets 0$;
            \FOR{$ epoch =1 \ \text{to} \  E $}
                \STATE Get \textit{NC} masks via Eq.~(\ref{eq: mask});
                \STATE Train NCGCN via Eq.~(\ref{eq: layer}) - (\ref{eq: loss});
                \STATE Get validation accuracy $Acc_\text{val}$;
                \IF {$Acc_\text{val} > Acc_\text{max}$}
                    \STATE $ Acc_\text{max} \gets Acc_\text{val} $;
                    \STATE Get pseudo labels $\hat{\boldsymbol{Y}}$ for $\forall v_i \in V$;
                    \STATE Update $\textit{NC}_i$ for $\forall v_i \in V$ by Eq.~(\ref{eq:NC});
                \ENDIF
                \IF {$Acc_\text{val} \leq Acc_\text{max}$ for $S$ epochs}
                \STATE Break;
                \ENDIF 
            \ENDFOR
            \STATE Get test accuracy $Acc_\text{test}$;
    \RETURN {$Acc_\text{test}$.}
    \end{algorithmic} 
\end{algorithm}

\subsection{Neighborhood Confusion-guided Message Passing}
We transform GCN with two separation strategies: 
1) the separation of transformation weights at the first layer enables learners in each group to focus on specific optimization processes and eliminates noise from others;
2) the separation of message passing at the second layer prevents the outputs of different learners from being inappropriately mixed.
Such separation designs bring two additional conveniences: 
1) raw features can be incorporated with group-specific scalar weights;
2) the model complexity of different learners can be controlled by different dropout rates.
These two kinds of separations can be easily achieved by deforming the GCN formula. 
Specifically, we formalize the NCGCN using the \nc mask as follows:
\begin{equation}\label{eq: layer}
    \begin{aligned}
       \text{Layer 1:} \quad \boldsymbol{H}_1^{s} & =\sigma(\operatorname{norm}(\boldsymbol{M}^{s}\boldsymbol{A}) \cdot \boldsymbol{X}\boldsymbol{W}_1^{s})\\
       \text{Layer 2:} \quad \boldsymbol{H}_2^{s} & =\sigma(\operatorname{norm}(\boldsymbol{A}\boldsymbol{M}^{s}) \cdot \boldsymbol{H}_1^{s}\boldsymbol{W}_2^{s})
    \end{aligned}
\end{equation}
where $s \in \{\text{low}, \text{high} \}$ indicates the learning channels for different node groups, $ \boldsymbol{W}_1^{s} \in \mathbb{R}^{d\times d'}$ and $ \boldsymbol{W}_2^{s} \in \mathbb{R}^{d'\times d'} $ are the weight matrices, $ \sigma(\cdot) $ is the activation function, $ d' $ is the dimension of hidden layers.
In the first layer, left multiplication by $\boldsymbol{M}^{s}$ implies performing row masking or \textit{target masking} on $\boldsymbol{A}$, which maintains the rows of the corresponding groups in $\boldsymbol{A}$, and further realizes the weight separation, i.e., the weight $\boldsymbol{W}_1$ is not shared between low-\nc and high-\nc nodes.
In doing so, $\boldsymbol{H}_1$ will aggregate all the first-order information and keep zero in the masked rows on each subsequent layer.
In the second layer, right multiplication by $\boldsymbol{M}^{s}$ implies performing column masking or \textit{source masking} on $\boldsymbol{A}$,
which can filter the biased information from different groups, 
so that the target node will only aggregate information from the same group of neighbors in the neighborhood:
\begin{equation*}
    target \leftarrow source \  \in \{low\leftarrow low, \  high\leftarrow high  \}
\end{equation*}
The above two learning channels $s \in \{\text{low}, \text{high} \}$ are used to characterize low-\nc and high-\nc nodes, respectively. 
Considering the importance of raw features, we introduce a linear transformation parameterized by $\boldsymbol{W}_x \in \mathbb{R}^{d\times d'}$ to recover them. Then we utilize learnable scalar weight $\alpha^s$ to combine the two outputs for each channel by:
\begin{equation}\label{eq: raw mix}
    \boldsymbol{H}_x =\boldsymbol{X}\boldsymbol{W}_x, \quad
    \boldsymbol{H}_{o}^s = \alpha^s\cdot\boldsymbol{H}_{2}^{s}+(1-\alpha^s)\cdot\boldsymbol{H}_x
\end{equation}
Finally, we derive the soft assignment prediction $ \boldsymbol{B} \in \mathbb{R}^{n\times C}$ through a MLP parameterized by $\boldsymbol{W}_o \in \mathbb{R}^{d'\times C}$ as follows:
\begin{equation}\label{eq:readout}
    \boldsymbol{B} = \operatorname{softmax}\left(\left(\boldsymbol{H}_{o}^{\text{low}}+\boldsymbol{H}_{o}^{\text{high}}\right) \cdot \boldsymbol{W}_o\right)
\end{equation}

\subsection{Model Training}\label{sec: training}
We employ the cross-entropy as the classification loss:
\begin{equation}\label{eq: loss}
    \mathcal{L}= - \operatorname{trace}(\boldsymbol{Y}_\text{train}^\top \cdot \log\boldsymbol{B})
\end{equation}
where $\operatorname{trace}(\cdot) $ means the sum of the diagonal elements of the matrix.
The pseudo code implementation is shown in Algorithm~\ref{alg:NCgcn}.
During model training, once the validation accuracy reaches a new high, the \nc values and masks of all nodes will be recalculated and updated, and used in subsequent epochs. Unlike other metrics, the low computational complexity of \nc metric allows it to be calculated and updated in real time during model training. The two processes of metric estimation and model inference will be alternately optimized to achieve better node classification.

\begin{table*}[!htb]
    \renewcommand\arraystretch{1.1}
    \centering
    \caption{Results on real-world benchmarks: Mean Test Accuracy (\%) $\pm$ Standard Deviation. Boldface letters mark the best results while underlined letters indicate the second best. Accuracies that rank first on the public leaderboard are marked with $^\dagger$. Results below dashed line represent ablation and do not participate in the comparison. Separ. is abbreviated from separation.}
    \label{tab:main}
    \resizebox{\textwidth}{!}{
    \begin{tabular}{c|ccccc|ccccc} 
    \hline\hline
    \diagbox{Method}{Dataset}       & \textbf{Pubmed}                & \textbf{Photo}                 & \textbf{Computers}             & \begin{tabular}[c]{@{}c@{}}\textbf{Coauthor} \\\textbf{CS}\end{tabular} & \begin{tabular}[c]{@{}c@{}}\textbf{Coauthor} \\\textbf{Physics}\end{tabular} & \textbf{Cora Full}             & \textbf{Actor}                 & \textbf{Penn94}                & \begin{tabular}[c]{@{}c@{}}\textbf{Chameleon}\\\textbf{Filtered}\end{tabular} & \begin{tabular}[c]{@{}c@{}}\textbf{Squirrel}\\\textbf{Filtered}\end{tabular}  \\ 
    \hline 
    \#Classes                                 & 3                                   & 8                                  & 10                                     & 15                                                                                          & 5                                                                                                & 70                                     & 5                                  & 2                                   & 5                                                                                                 & 5                                                                                                 \\
    \#Features                                & 500                                 & 745                                & 767                                    & 6805                                                                                        & 8415                                                                                             & 8710                                   & 932                                & 4814                                & 2325                                                                                              & 2089                                                                                              \\
    \#Nodes                                   & 19717                               & 7650                               & 13752                                  & 18333                                                                                       & 34493                                                                                            & 19793                                  & 7600                               & 41554                               & 864                                                                                               & 2205                                                                                              \\
    \#Edges                                   & 44324                               & 119081                             & 245861                                 & 81894                                                                                       & 247962                                                                                           & 63421                                  & 26659                              & 1362229                             & 7754                                                                                              & 46557                                                                                             \\ 
    \textit{NH}                             & 0.87                                & 0.86                               & 0.82                                   & 0.86                                                                                        & 0.93                                                                                             & 0.69                                   & 0.62                               & 0.56                                & 0.33                                                                                              & 0.27                                                                                              \\
    \textit{NC} ($k=1$)                     & 0.14                                & 0.07                               & 0.08                                   & 0.06                                                                                        & 0.05                                                                                             & 0.09                                   & 0.32                               & 0.68                                & 0.51                                                                                              & 0.60                                                                                              \\
    \textit{NC} ($k=2$)                     & 0.20                                & 0.18                               & 0.20                                   & 0.13                                                                                        & 0.10                                                                                             & 0.16                                   & 0.43                               & 0.90                                & 0.58                                                                                              & 0.72                                                                                              \\
    \hline
    MLP                             & 86.86 $\pm$ 0.51          & 91.28 $\pm$ 0.63          & 84.35 $\pm$ 1.02          & 95.22 $\pm$ 0.17          & 95.63 $\pm$ 0.24          & 60.60 $\pm$ 0.77          & 36.82 $\pm$ 1.19          & 75.34 $\pm$ 0.64          & 31.99 $\pm$ 4.11           & 44.61 $\pm$ 4.23           \\
    GCN                             & 88.08 $\pm$ 0.50          & 93.94 $\pm$ 0.47          & 89.72 $\pm$ 0.43          & 94.24 $\pm$ 0.27          & 96.31 $\pm$ 0.15          & 70.42 $\pm$ 0.36          & 33.15 $\pm$ 1.75          & 76.55 $\pm$ 0.57          & 42.63 $\pm$ 3.64           & 47.83 $\pm$ 2.90           \\
     SAGE                             & 88.73 $\pm$ 0.67& 93.32 $\pm$ 0.45& 90.01 $\pm$ 0.77& 94.35 $\pm$ 0.58& 97.39 $\pm$ 0.18& 70.51 $\pm$ 0.60& 36.37 $\pm$ 0.21& OOM& 45.15 $\pm$ 3.99& 51.64 $\pm$ 2.68   \\
    GAT                             & 88.74 $\pm$ 0.52& 94.21 $\pm$ 0.57          & 89.74 $\pm$ 0.47          & 94.35 $\pm$ 0.25          & 96.63 $\pm$ 0.25          & 70.02 $\pm$ 0.57          & 36.55 $\pm$ 1.28          & 77.68 $\pm$ 2.37          & 45.09 $\pm$ 4.34           & 44.11 $\pm$ 5.59           \\
    \hline
    H2GCN                           & 89.43 $\pm$ 0.48          & \underline{95.48 $\pm$ 0.43}  & 89.68 $\pm$ 1.02          & 95.85 $\pm$ 0.33          & 97.64 $\pm$ 0.50          & 71.01 $\pm$ 0.42          & 32.62 $\pm$ 2.91          & 83.20 $\pm$ 0.55          & 45.03 $\pm$ 6.46           & 41.47 $\pm$ 1.55           \\
    GPRGNN                          & 90.46 $\pm$ 0.75  & 93.97 $\pm$ 0.67          & 87.92 $\pm$ 1.96          & 95.74 $\pm$ 0.31          & 97.77 $\pm$ 0.15  & 71.37 $\pm$ 0.48          & 39.22 $\pm$ 1.17          & \underline{84.55 $\pm$ 0.54}  & 45.15 $\pm$ 4.95           & 45.85 $\pm$ 2.33           \\
    FAGCN                           & 89.16 $\pm$ 0.67          & 94.89 $\pm$ 0.48          & 87.61 $\pm$ 1.17          & 95.94 $\pm$ 0.56          & 97.37 $\pm$ 0.33          & 69.10 $\pm$ 1.17          & 40.90 $\pm$ 1.12  & 76.97 $\pm$ 0.69          & 45.20 $\pm$ 4.45   & 53.28 $\pm$ 1.28   \\
    ACM-GCN                         & 90.40 $\pm$ 0.29          & 95.04 $\pm$ 0.44          & 90.31 $\pm$ 0.39  & 96.10 $\pm$ 0.22  & 97.09 $\pm$ 0.20          & 72.19 $\pm$ 0.36 & 39.09 $\pm$ 1.42          & 82.98 $\pm$ 0.63          & 43.74 $\pm$ 4.03           & 39.64 $\pm$ 7.84           \\ 
    \hline
    CPGNN                           & 87.54 $\pm$ 1.72          & 90.49 $\pm$ 2.43         & 83.50 $\pm$ 2.23          & 95.42 $\pm$ 0.24          & 97.61 $\pm$ 0.48          & 68.47 $\pm$ 1.02          & 33.52 $\pm$ 1.71          & OOM                   & 39.18 $\pm$ 1.28           & 28.32 $\pm$ 2.11           \\
    WRGAT                           & 89.13 $\pm$ 0.53          & 92.23 $\pm$ 0.64          & 86.13 $\pm$ 0.76          & 95.08 $\pm$ 0.28          & OOM                   & 68.67 $\pm$ 0.37          & 37.91 $\pm$ 1.63          & OOM                   & 42.46 $\pm$ 3.71           & 45.47 $\pm$ 3.56           \\
    GBKSage                         & 85.49 $\pm$ 0.61          & 81.61 $\pm$ 8.05          & 81.20 $\pm$ 5.62          & 95.22 $\pm$ 0.28          & OOM                   & 69.19 $\pm$ 0.47          & 37.27 $\pm$ 0.98          & OOM                   & 45.32 $\pm$ 4.61        & 45.41 $\pm$ 3.19           \\
    HOG                             & 85.25 $\pm$ 0.77          & 90.88 $\pm$ 0.68          & 82.07 $\pm$ 1.31          & 95.94 $\pm$ 0.16          & OOM                   & 60.96 $\pm$ 0.86          & 38.00 $\pm$ 1.72          & OOM                   & 41.87 $\pm$ 5.94           & 39.92 $\pm$ 2.38           \\
    GloGNN                          & 84.87 $\pm$ 0.61          & 92.41 $\pm$ 0.72          & 86.05 $\pm$ 1.71          & 91.54 $\pm$ 5.14          & 97.54 $\pm$ 0.53          & 67.04 $\pm$ 0.75          & 32.88 $\pm$ 2.03          & 83.07 $\pm$ 1.35          & 44.39 $\pm$ 3.99           & 49.07 $\pm$ 8.68           \\
    SNGNN                           & 86.85 $\pm$ 0.62          & 90.79 $\pm$ 2.59          & 85.30 $\pm$ 1.02          & 95.52 $\pm$ 0.28          & 96.43 $\pm$ 0.10          & 71.11 $\pm$ 0.31          & 38.68 $\pm$ 0.78          & 73.87 $\pm$ 1.01          & 41.17 $\pm$ 3.61           & 38.27 $\pm$ 2.62           \\ 
    CAGNN                          & 87.54 $\pm$ 0.67& 92.75 $\pm$ 1.29& 89.59 $\pm$ 0.63& 94.50 $\pm$ 0.81& 97.93 $\pm$ 0.30& 69.88 $\pm$ 0.53& 39.83 $\pm$ 0.85& 79.49 $\pm$ 2.00& 43.86 $\pm$ 2.98& 53.28 $\pm$ 2.26\\ 
    \hline
    NCGCN                           & \textbf{91.64 $\pm$ 0.53}$^\dagger$ & 95.45 $\pm$ 0.45 & \textbf{90.81 $\pm$ 0.46} & \textbf{96.64 $\pm$ 0.29}$^\dagger$ & \underline{98.63 $\pm$ 0.24} & \textbf{73.42 $\pm$ 0.58}$^\dagger$  & \underline{43.16 $\pm$ 1.32} & \textbf{84.74 $\pm$ 0.28} & \underline{46.84 $\pm$ 4.36} & \underline{54.15 $\pm$ 1.47}  \\ 
      NCSAGE                           &   \underline{91.55 $\pm$ 0.38} & \textbf{95.93 $\pm$ 0.36}$^\dagger$ & \underline{90.43 $\pm$ 0.72} & \underline{96.48 $\pm$ 0.25} & \textbf{98.69 $\pm$ 0.26}$^\dagger$ & \underline{72.58 $\pm$ 0.65} & \textbf{43.89 $\pm$ 1.33}$^\dagger$ & 81.77 $\pm$ 0.71 & \textbf{48.07 $\pm$ 3.47} & \textbf{54.42 $\pm$ 1.41} \\ 
    \hdashline
    \textit{w/o} Separ.             &   90.73 $\pm$ 0.76 & 94.22 $\pm$ 0.72 & 88.45 $\pm$ 0.76 & 94.43 $\pm$ 0.53 & 97.82 $\pm$ 0.24 & 69.70 $\pm$ 0.70 & 40.01 $\pm$ 1.68 & 79.07 $\pm$ 1.32 & 36.49 $\pm$ 4.13 & 53.89 $\pm$ 1.37 \\
    \textit{w/o} Mess. Separ.     & 90.93 $\pm$ 0.49          & 93.69 $\pm$ 0.61          & 87.10 $\pm$ 0.56          & 95.21 $\pm$ 0.38          & 96.83 $\pm$ 0.67          & 67.98 $\pm$ 0.60          & 39.97 $\pm$ 2.16          & 76.30 $\pm$ 0.40          & 32.57 $\pm$ 3.18           & 36.72 $\pm$ 4.08           \\
    \textit{NH} Separ.   & 90.88 $\pm$ 0.56          & 94.54 $\pm$ 0.42          & 89.71 $\pm$ 0.53          & 95.75 $\pm$ 0.41          & 98.06 $\pm$ 0.40          & 66.17 $\pm$ 0.73          & 39.03 $\pm$ 1.63          & 78.52 $\pm$ 0.51          & 39.24 $\pm$ 6.64           & 31.81 $\pm$ 5.08           \\
    \hdashline
    NCGCN ($k=3$)                  & 91.17 $\pm$ 0.62  & 95.06 $\pm$ 0.59 & 90.42 $\pm$ 0.58 & 94.86 $\pm$ 0.44 & 97.63 $\pm$ 0.28 & 56.33 $\pm$ 0.70 & 42.20 $\pm$ 1.52 & 76.81 $\pm$ 0.67 & 35.38 $\pm$ 2.05 & 35.26 $\pm$ 5.40 \\
    \hline\hline
    \end{tabular}}
\end{table*} 

\subsection{Plug-in Property and Efficiency}\label{sec: plugin}
While we focus on separate learning research on the classic message passing architecture of GCN, it's notable that our \nc-guided framework possesses a plug-in property, i.e., it can be combined with any other message passing neural networks by simply replacing the backbone model. To show the plug-in property of our framework, we further replaced GCN with another message passing neural network, SAGE~\cite{hamilton2017inductive}, forming the NCSAGE model. Its performance will be demonstrated in the experimental section.

In addition, our framework introduces three key optimizations at design time, which include:
1) constructing a hash dictionary to efficiently index the $k$-hop neighbors of a node, enabling the estimation of \nc values to operate with linear complexity; 
2) computing the two separated channels in parallel;
3) using threshold-based separated learning to make the learner of each channel sparser than the corresponding backbone in terms of parameters. 
These architectural features avoid imposing excessive computational overhead, and ultimately ensure good computational efficiency.

\section{Evaluations}
\subsection{Experimental Settings}
\subsubsection{Datasets}
We use 10 real-world benchmarks from~\cite{chien2020adaptive,shchur2018pitfalls,lim2021linkx}, among which the first five (\texttt{Pubmed}, \texttt{Photo}, \texttt{Computers}, \texttt{Coauthor CS} and \texttt{Coauthor Physics}) are homophilous and the last five (\texttt{Cora Full}, \texttt{Actor}, \texttt{Penn94}, \texttt{Chameleon} and \texttt{Squirrel}) are heterophilous. Note that such division is based on the criterion of $\textit{NH}=0.7$.
We adopt de-duplicated version of the last two, namely \texttt{Chameleon Filtered} and \texttt{Squirrel Filtered}, as provided in~\cite{critical}.
For all datasets, we adopt dense data splitting as in~\cite{chien2020adaptive}, i.e., randomly split them into training, validation and testing sets with a proportion of 60\%/20\%/20\%. 
We provide more details for these datasets in Appendix~\ref{app:datasets}.

\begin{figure*}[!ht]
    \centering
    \includegraphics[width =\textwidth]{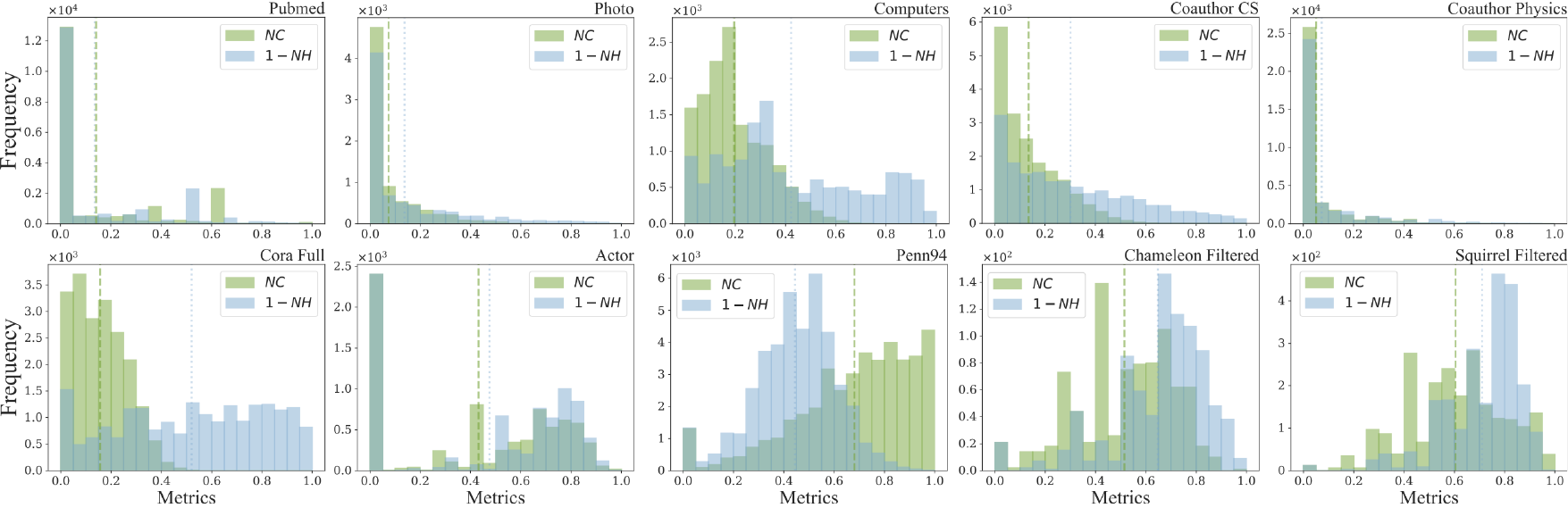}
    \caption{Distributions of nodes guided by \nh and \nc, with dashed lines representing the respective averages. Note that using 1-\nh instead of \nh facilitates comparison between these two metrics. The hops of \nc adhere to the optimal value on NCGCN for each dataset during hyperparameter tuning.}
    \label{fig:dist}
\end{figure*}

\subsubsection{Competitors}
We compare our methods with a total of 15 competitors in three categories, including classical methods such as two-layer MLP, GCN~\cite{kipf2016semi}, SAGE~\cite{hamilton2017inductive} and GAT~\cite{velivckovic2017gat}, spectral methods such as H2GCN~\cite{zhu2020h2gcn}, GPRGNN~\cite{chien2020adaptive}, FAGCN~\cite{bo2021fagcn} and ACM-GCN~\cite{luan2022acm}, spatial methods such as CPGNN\cite{zhu2021cpgnn}, WRGAT~\cite{suresh2021wrgat}, GBKSage~\cite{du2022gbk}, HOG~\cite{wang2022hog}, GloGNN~\cite{li2022glognn}, CAGNN~\cite{cagnn} and SNGNN~\cite{zou2023similarity}. The methods are the latest and tailor-made for the heterophily problem. We provide more details for these competitors in Appendix~\ref{app:baselines}.

\subsubsection{Setup}
To ensure the stability and reproducibility of the results, we use ten consecutive seeds to fix the data splitting and model initialization. For all methods, we set the maximum epochs to 500 with 100 epochs patience for early stopping, the hidden dimension to 512, and the optimizer to Adam~\cite{kingma2014adam}. For common parameters like learning rate, weight decay and dropout rate, we search in the same parameter space if their codes are publicly available. 
The key parameters of NCGCN/NCSAGE like hop $k$ and threshold $T$ will be searched in their respective parameter spaces $k\in\{1, 2\}$ and $T\in\{0.3,0.4,0.5,0.6,0.7\}$. 
Detailed parameter settings are reported in Appendix~\ref{app:parameter}.

\subsection{Analysis}
To clearly organize this section, we answer six \textbf{R}esearch \textbf{Q}uestions (\textbf{RQ}) and demonstrate our arguments by extended experiments. 
For simplicity, we only discuss the validity of our framework on NCGCN in \emph{RQ2} - \emph{RQ6}.

\subsubsection{\textbf{RQ1: Does our framework outperforms baselines and its backbone across homophilous and heterophilous datasets?}}
Table~\ref{tab:main} reports the node classification results of all methods on ten benchmarks, from which we can draw the following conclusions: 
1) Among all baselines, spectral methods show a dominant performance advantage on the heterophily problem. 
However, although they achieve relatively high overall performance, they lack a certain degree of universality, e.g., GPRGNN does not perform well on \texttt{Photo} and \texttt{Computer}. This is mainly due to the fact that the high-order prior of these methods is not applicable to all datasets, and the weight sharing hinders further performance improvement. 
2) Spatial methods usually have complex architectural designs and large hyperparameter spaces, leading to higher computational complexity as well as unstable and poor performance. For instance, WRGAT, GBKSage and HOG are out of memory (OOM) on \texttt{Coauthor Physics} and \texttt{Penn94}. 
3) As a spatial method, our framework achieves the best overall performance on two types of datasets without being out of memory (OOM), and reaches first place on the public leaderboard in six well-explored benchmarks (refer to Appendix~\ref{app: public-leaderboard}), indicating its performance advantage.
Additionally, our framework shows an average performance improvement of 8.04\% and 4.54\% on their respective backbones (GCN and SAGE), confirming its general plug-in property. 
Notably, the sparsification of the adjacency matrix by the \nc-mask allows our NCSAGE model to overcome the OOM problem encountered by SAGE on \texttt{Penn94}.
All these phenomena highlight the effectiveness of our framework in alleviating heterophily problem, especially the superiority of our proposed \nc metric and priors.

\begin{figure*}[!ht]
    \centering
    \includegraphics[width =\linewidth]{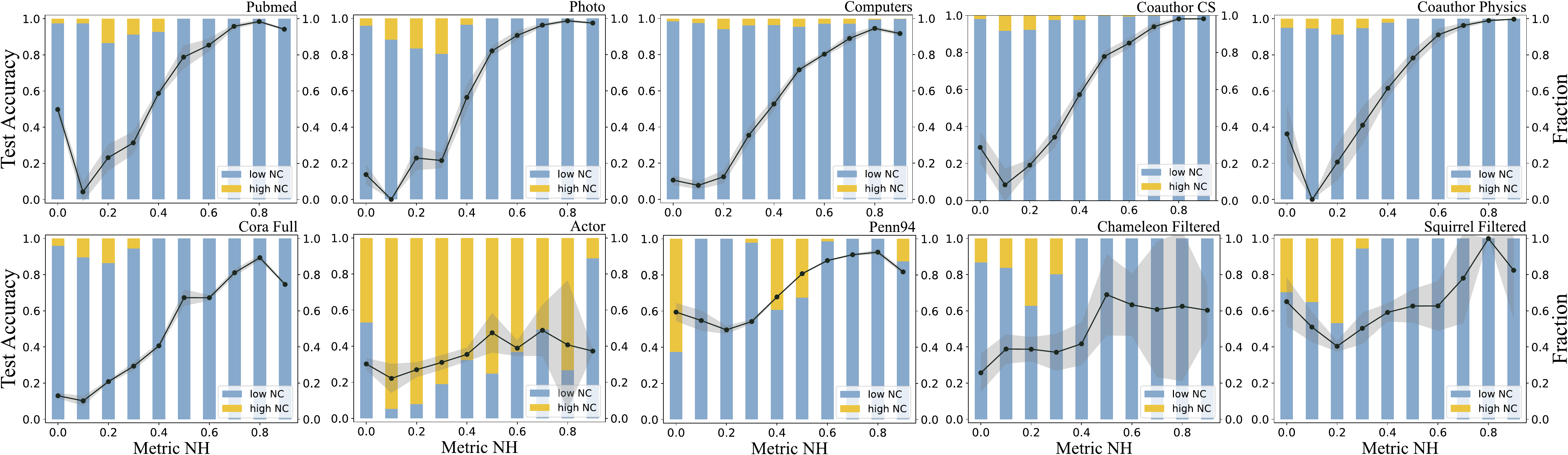}
    \caption{More results of Fig.~\ref{fig:fraction}: The test accuracy of GCN in different node groups varies with \nh value and proportion of high-\nc nodes.}
    \label{fig:nh-acc}
\end{figure*}
\begin{figure*}[!ht]
    \centering
    \includegraphics[width =\linewidth]{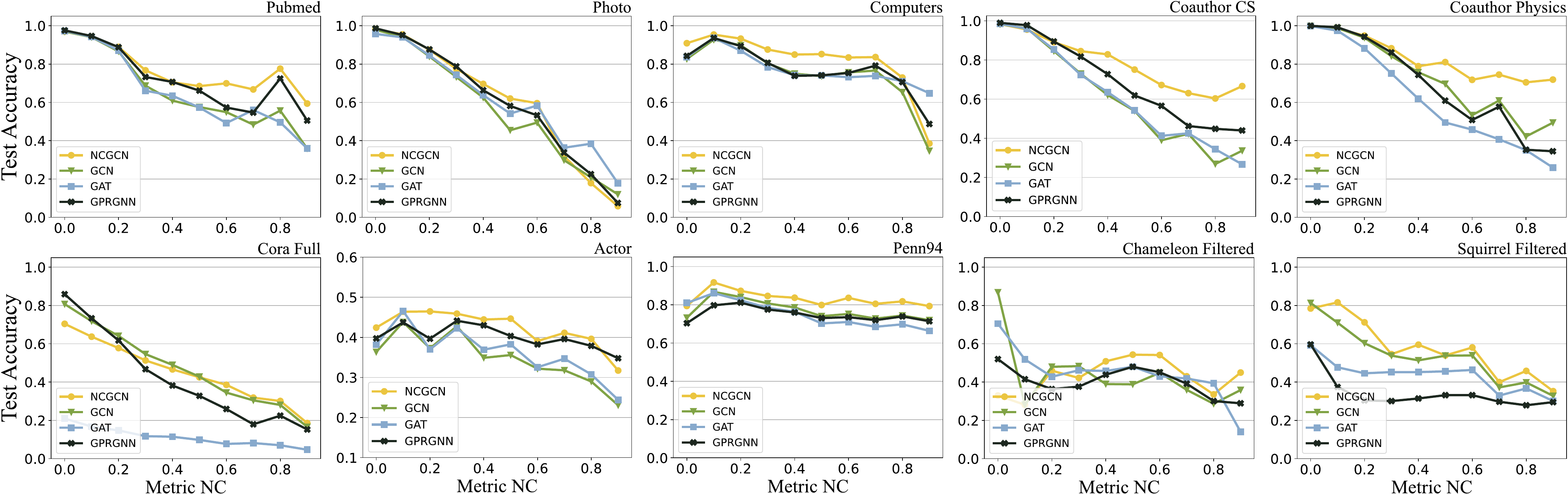}
    \caption{Test accuracies of different models under different \textit{NC} groups.}
    \label{fig:nc-acc}
    \end{figure*}

\subsubsection{\textbf{RQ2: How do the key designs in NCGCN contribute?}}
Our framework features three key designs: 
1) the separation of transformation weights in the first layer; 
2) the separation of message passing in the second layer; 
3) the guidance of separation by the \nc metric. 
To further investigate the effectiveness of these key designs, we conduct three ablation studies: 
1) removing separation strategies (target masking and source masking) and degenerating NCGCN to an adaptive mixture of GCN output and raw features (\textit{w/o} Separ.); 
2) restoring inter-group message passing (source masking) in the second layer (\textit{w/o} Mess. Separ.); 
3) guiding separation using \nh metric and a threshold of 0.5 (\textit{NH} Separ.). 
The results of ablation studies are shown below the dashed line in Table~\ref{tab:main}.
By comparing \textit{w/o} Separ. and \textit{w/o} Mess. Separ. with NCGCN, we conclude that introducing separation strategies enable more high-\nc nodes to benefit from a more appropriate feature learning process, especially in heterophilous graphs where the separation strategies yield higher gains.
Additionally, \textit{NH} Separ. outperforms \textit{w/o} Separ. on homophilous graphs but fails on heterophilous ones, indicating that \textit{NH} metric has limited generalizability on heterophilous graphs, reinforcing the superiority of our \textit{NC}-guided separation strategy. 
In conjunction with the node distributions under different metrics in Fig.~\ref{fig:dist}, it is evident that \nc and \nh hold different insights on the confusion of nodes in the graph. Our \nc metric facilitates a more effective re-adjustment of node distribution, particularly in heterophilous graphs.
In summary, the performance gap between ablation models and our NCGCN highlights the effectiveness of the three key designs.

\subsubsection{\textbf{RQ3: Whether the \nc metric can measure the distinguishability of nodes?}}
Before analyzing the interpretability of \nc metric, we first review the perplexing phenomenon exhibited by \nh metric in Fig.~\ref{fig:nh-acc}. We can see that the grouped test accuracy curves based on \nh metric consistently present a `ticking' trend, especially where the node groups with extremely low homophily show high test accuracy. It cannot be explained well by the definition of \nh metric.
However, there seems to be a certain negative correlation between test accuracy and the proportion of high-\nc nodes, offering an insight into this phenomenon's interpretation. 

We further group nodes based on \nc metric and record the grouped test accuracy curves of different models, as shown in Fig.~\ref{fig:nc-acc}. Firstly, we note that in most cases, the grouped test accuracy curves monotonically decrease as \nc value increases. This common trend can be reasonably explained as follows.
When nodes have neighbors with multiple label types, they exhibit higher \nc values according to the definition of neighborhood confusion. In such cases, nodes receive more heterophilous information during message passing, which can introduce noise, thereby reducing their distinguishability. This decreased distinguishability might present greater challenges in classifying these nodes, consequently affecting test accuracy. Therefore, we can conclude that \nc metric can, to some extent, measure the distinguishability of nodes, with nodes having higher \nc values being less distinguishable.

Furthermore, it is observable that compared to some baseline methods, our separation learning framework typically improves the classification performance of different \nc groups more effectively. In a weight-sharing learning process, the idea of simplifying the model conflicts with the eliminating biases. This conflict causes interference between the different patterns of low-\nc and high-\nc nodes, forcing the model to make a trade-off between fairness and accuracy. In contrast, our separation learning framework breaks the assumption of weight sharing, offering customized learners for both low-\nc and high-\nc nodes, thus holding the potential for debiasing.

\begin{figure*}[!ht]
    \centering
    \includegraphics[width = \linewidth]{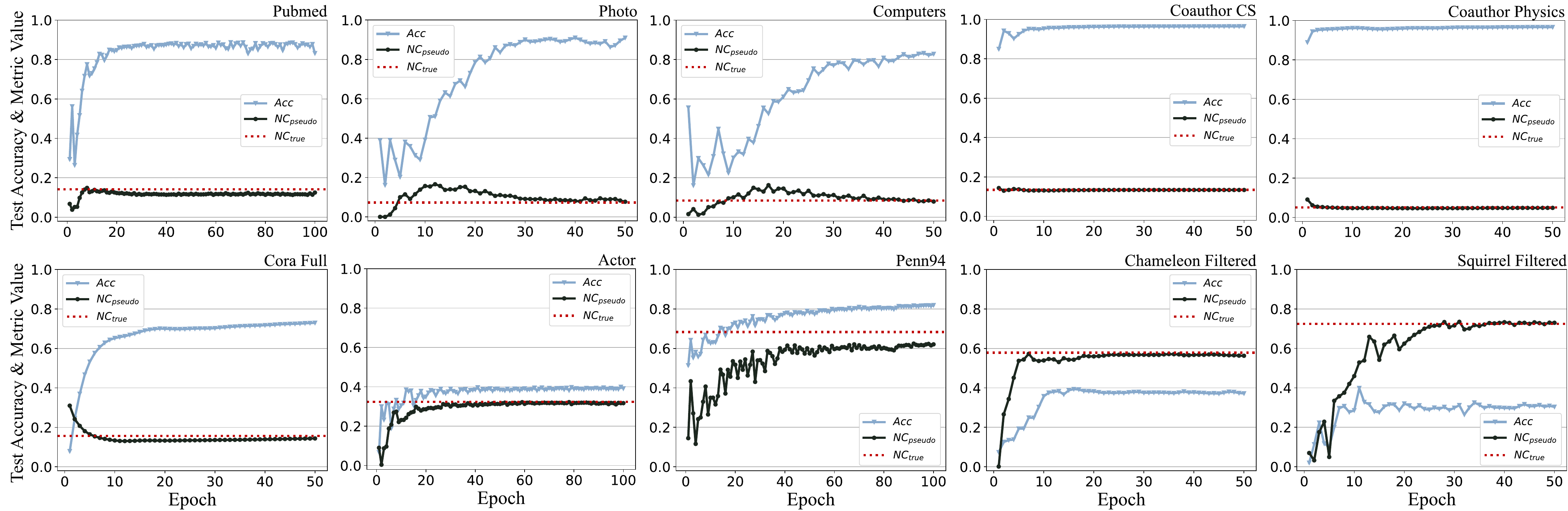}
    \caption{Illustration of interplay between metric estimation and model inference during training.}
    \label{fig: acc-curve}
\end{figure*}

\begin{figure*}[!ht]
    \centering
    \includegraphics[width = \linewidth]{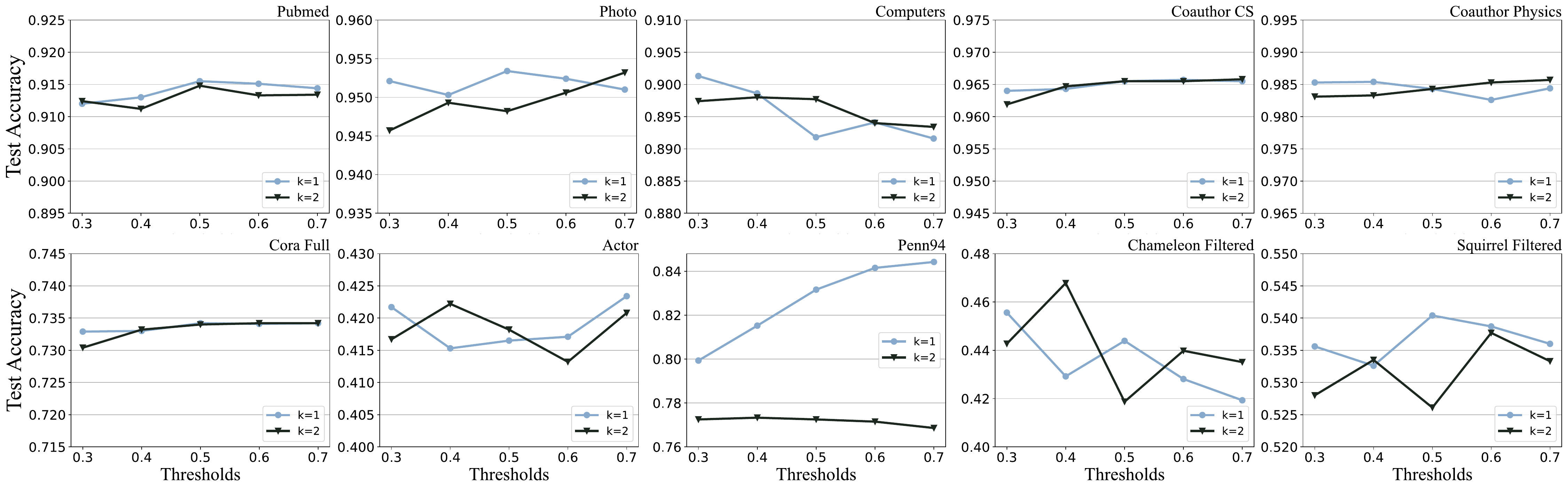}
    \caption{Sensitivity of NCGCN with respect to $k$ hops of \textit{NC} and the threshold $T$.}\label{fig:hyper}
\end{figure*}

\subsubsection{\textbf{RQ4: Whether metric estimation and model inference complement each other?}}
Our framework relies on predicted pseudo-labels to estimate \nc values, which subsequently guide the separate learning process. 
On the one hand, the accuracy of the pseudo-labels predicted by the model, i.e., the effectiveness of model inference, will impact the accuracy of metric estimation.
On the other hand, the accuracy of metric estimation will impact the \nc mask and subsequently impact node separation results, finally impacting both model training and inference. Therefore, we further analyze the interplay between metric estimation and model inference. We first initialize $\boldsymbol{M}^\text{low}=\boldsymbol{I}$, meaning all nodes are initially assigned to the low-\nc group, and then commence model training.
Fig.~\ref{fig: acc-curve} shows the curves of average test accuracy and the average pseudo \nc value across 10 runs.
Note that we use real labels to calculate the ground-truth \nc value at the global level (red dashed line), and use the predicted pseudo-labels to estimate the pseudo \nc value (black curve).
In the early stages of training, we observe continuous fluctuations in the model's test accuracy and the estimated \nc value. With the increase of training epochs, the \nc value estimated by the model gradually align more closely with the ground-truth \nc value, and the model's test accuracy progressively improves. Both aspects tend to stabilize after a certain number of training epochs, particularly the metric estimation results which, in most cases, converge to the ground-truth. This phenomenon suggests that the processes of metric estimation and model inference are synergistically optimized during training, complementing each other and enhancing the overall performance.

Additionally, it's noteworthy that our framework is initialized with no high-\nc nodes, but the estimated \nc values can eventually converge to the ground truth \nc, suggesting that the node separation task (i.e., metric estimation) is easier to optimize than the node classification task (i.e., model inference). This is a significant advantage of our proposed metric and framework: even when the model cannot achieve high prediction performance, our metric can still robustly separate the nodes in a reasonable manner.

\begin{figure*}[!ht]
\centering
        \includegraphics[width = \linewidth]{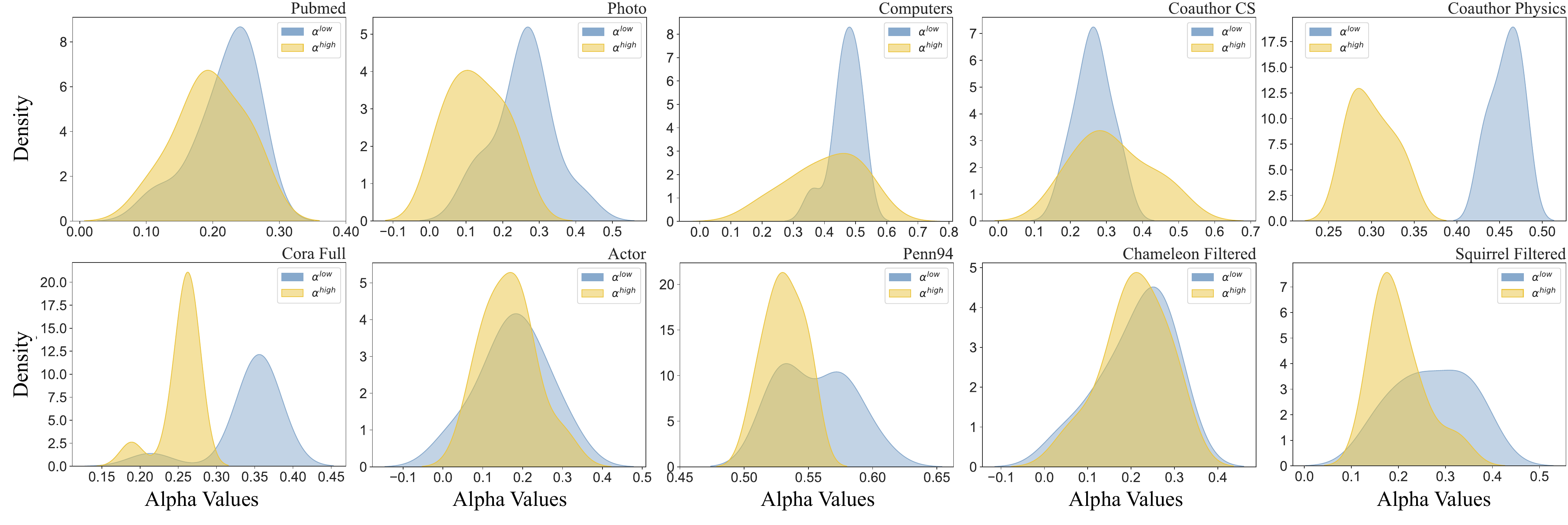}
    \caption{The kernel density estimates of weights $\alpha^{s}$ (See Eq.~(\ref{eq: raw mix})) for aggregated features and raw features.}
    \label{fig:scale-weight}
\end{figure*}

\subsubsection{\textbf{RQ5: How do the two hyperparameters of hops $k$ and threshold $T$ impact NCGCN?}}
Metric computation and node separation in our framework are controlled by the two key parameters, hops $k$ and threshold $T$. To further investigate the impact of these hyperparameters on NCGCN, we fix the other parameters to their optimal settings and plot the test accuracy curves under varying $k$ and $T$, as shown in Fig.~\ref{fig:hyper}.

\begin{figure}[!ht]
    \centering
    \includegraphics[width = 0.8\linewidth]{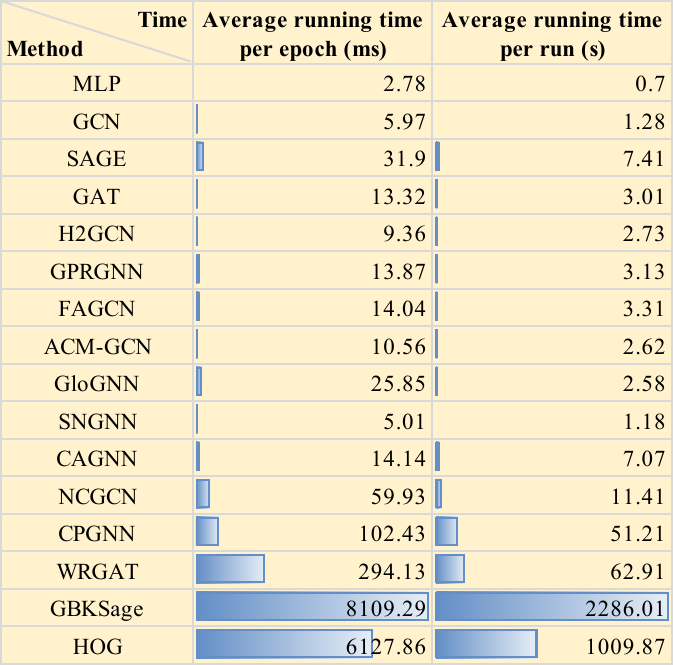}
    \caption{Average running time per epoch (ms) and per run (s).}
    \label{fig: time}
\end{figure}

When we fix $T$, we observe that NCGCN exhibits relatively small performance fluctuations with varying $k$ on the five homophilous graphs and \texttt{Cora Full}. This is because these datasets have relatively stable label distributions, and the \nc values computed under different parameter settings are more consistent, resulting in similar node groupings. Consequently, NCGCN is more robust to changes in the hop parameter. In contrast, the neighborhood of different hops in heterophilous graphs will exhibit more diverse label distributions, resulting in significantly distinct node groupings. As a result, NCGCN demonstrates heightened sensitivity to parameter changes. For large-scale heterophilous graphs like \texttt{Penn94}, the 1-hop receptive field already includes a sufficient number of neighbors. The main reason for its poor performance in the 2-hop setting lies in the ineffective utilization of 2-hop information by the target nodes to evaluate their own state.

When we fix $k$, we observe that NCGCN exhibits relatively small performance fluctuations with varying $T$ on the five homophilous graphs and \texttt{Cora Full}. This is because these datasets typically have a large proportion of low-\nc nodes, as shown in Fig.~\ref{fig:dist}, and as the threshold increases, the node groupings remain stable, leading to consistent model performance. For small-scale heterophilous graphs like \texttt{Actor}, \texttt{Squirrel Filtered}, and \texttt{Chameleon Filtered}, NCGCN is more sensitive to threshold changes due to the more discrete distribution of nodes across different \nc values. For larger heterophilous graphs like \texttt{Penn94}, the distribution of nodes with different \nc values tends to increase with higher \nc values. As the threshold increases, the division between low-\nc and high-\nc nodes becomes more balanced, which gradually improves the performance of NCGCN ($k=1$).

\subsubsection{{\textbf{RQ6: Whether separate learning setting exhibit different feature fusion preferences in different node groups?}}}\label{sec: scale-weight}
In Eq.~(\ref{eq: raw mix}), we introduce learnable scalar weight $\alpha^s$ for each learner to fuse aggregated features with raw features. To further explore whether learners acting on different node groups have distinct preferences in feature fusion, we plot the kernel density estimates of $\alpha^s$ for the low- and high-\nc groups across ten runs, as shown in Fig.~\ref{fig:scale-weight}. It is evident that the peak of the weight distribution for the low-\nc group is right-skewed compared to that of the high-\nc group. This implies that the learner in the low-\nc group assigns greater weight to aggregated features and less to raw features during feature fusion, while the opposite is true for the high-\nc group. This phenomenon aligns with intuition and can be reasonably explained as follows. Nodes in the high-\nc group have neighborhoods with more complex label distributions, and the aggregated features obtained through message passing contain more heterophilous noise. Therefore, the learner for the high-\nc group tend to rely more on the nodes' raw features for representation, resulting in relatively smaller $\alpha^\text{high}$. Conversely, nodes in the low-\nc group have neighborhoods with purer label distributions, and the learner for this group effectively utilizes the features obtained through message passing for node representation, resulting in relatively larger $\alpha^\text{low}$.
In summary, the gap between the weights of two groups within one graph suggests that our separation strategy is necessary and effective.

\begin{table}
    \renewcommand\arraystretch{1.1}
    \centering
    \caption{Results on large-scale benchmarks with inductive learning.}
    \label{tab: large}
    \resizebox{0.8\linewidth}{!}{
    \begin{tabular}{c|c|c} 
    \hline\hline
    \multicolumn{1}{l|}{\diagbox{Method}{Dataset}} & \textbf{ogbn-arxiv}            & \textbf{Flickr}                 \\ 
    \hline
    \#Remark                                       & large, temporal       & large                  \\
    \#Classes                                      & 40                    & 7                      \\
    \#Features                                     & 128                   & 500                    \\ 
    \#Nodes                                        & 169343                & 89250                 \\ 
    \#Edges                                        & 1166243               & 334120                \\
    \textit{NH}                                    & 0.64                      & 0.32                       \\
    \textit{NC} ($k=1$)                            & 0.04                      & 0.37                       \\
    \textit{NC} ($k=2$)                            & 0.06                      & 0.57                       \\ 
    \hline
    MLP                                            & 55.39 $\pm$ 0.17          & 46.91 $\pm$ 0.09           \\
    GCN                                            & 55.87 $\pm$ 0.18          & 47.82 $\pm$ 0.11           \\
    SAGE                                           & 57.52 $\pm$ 0.29          & 47.87 $\pm$ 0.08           \\
    GAT                                            & 56.81 $\pm$ 0.23          & 47.28 $\pm$ 0.16           \\ 
    \hline
    H2GCN                                          & 55.23 $\pm$ 0.14          & 48.26 $\pm$ 0.11           \\
    GPRGNN                                         & 57.29 $\pm$ 0.28          & 48.03 $\pm$ 0.10           \\
    FAGCN                                          & 57.03 $\pm$ 0.27          & 48.12 $\pm$ 0.36           \\
    ACM-GCN                                        & 56.78 $\pm$ 0.19          & 47.54 $\pm$ 0.28           \\ 
    \hline
    CPGNN                                          & OOM                       & OOM                      \\
    WRGAT                                          & OOM                       & OOM                          \\
    GBKSAGE                                        & OOM                       & OOM                          \\
    HOG                                            & OOM                       & OOM                      \\
    GloGNN                                         & OOM                       & OOM                    \\
    SNGNN                                          & 56.21 $\pm$ 0.18          & 47.11 $\pm$ 0.13           \\
    CAGNN                                          & 50.34 $\pm$ 3.49          & 47.01 $\pm$ 0.10           \\ 
    \hline
    NCGCN                                          & \underline{58.29 $\pm$ 0.27}  & \textbf{48.43 $\pm$ 0.09}  \\
    NCSAGE                                         & \textbf{59.20 $\pm$ 0.32} & \underline{48.33 $\pm$ 0.13}   \\
    \hline\hline
    \end{tabular}}
    \end{table}

\subsubsection{\textbf{Efficiency Comparison}}\label{sec:efficiency}
To investigate the computational efficiency of different methods, we conduct a relatively fair comparison of actual time consumption on \texttt{Pubmed}, as shown in Fig.~\ref{fig: time}. For all methods, the number of \{hidden units, layers\} are set to \{64, 2\}, respectively.
The results demonstrate that our NCGCN outperforms partial spatial methods in terms of computational efficiency.

First, CPGNN, GBKSage and HOG employ two-phase training and soft rewiring techniques, resulting in significant time consumption.
WRGAT constructs a weighted multi-relational graph before training, rendering it equivalent to multi-channel GATs with reduced efficiency.
These spatial methods typically involve additional computational steps for adjusting message passing rules, such as rewiring, auxiliary tasks, or gating mechanisms, thereby increasing the computational burden.
In contrast, the NCGCN architecture is essentially a simple deformation of GCN, while also incorporates the three key designs mentioned in Sec.~\ref{sec: plugin}. Consequently, it avoids excessive computational overhead, ultimately ensuring good computational efficiency. Please refer to Appendix~\ref{app:eff} for more analysis.
Additionally, although some spatial methods consume less computation time, they have certain design flaws in their model architectures. For GloGNN, directly inputting the adjacency matrix into an MLP contradicts permutation invariance. For SNGNN, the approach of selecting the most similar neighbors seems to remain confined within the homophily assumption. CAGNN combines raw and aggregated features without considering label heterogeneity. These design flaws result in these methods falling short in terms of accuracy.

\subsubsection{\textbf{Performance Analysis on Large-scale Graphs with Inductive Learning}}
In this section, we further analyze the performance of various models on large-scale graphs with inductive learning, focusing on the \texttt{ogbn-arxiv}~\cite{wang2020microsoft} and \texttt{Flickr}~\cite{Zeng2020GraphSAINT} benchmarks. These two datasets present distinct challenges: \texttt{ogbn-arxiv} contains temporal information and a higher degree of homophily, while \texttt{Flickr} is more heterophilous, providing a diverse testing ground for models designed to address the heterophily issue in GNNs. As shown in Table~\ref{tab: large}, the proposed NCGCN and NCSAGE, which are guided by the \nc metric, deliver the best performance across both datasets. These results demonstrate the effectiveness of the \nc metric in separating nodes based on neighborhood label diversity, allowing the models to apply distinct learning processes to nodes with different levels of neighborhood confusion. This separation is especially crucial for handling the heterophilous nature of \texttt{Flickr} while maintaining high performance on the larger and more homophilous \texttt{ogbn-arxiv} dataset.
Additionally, it is important to note that several complex models, such as GBKSAGE and HOG, experience timeout and OOM errors, further demonstrating the challenge of scaling certain spatial methods to large-scale graphs. In contrast, NCGCN and NCSAGE handle these large-scale datasets efficiently, illustrating the scalability of the NC-guided separation framework in real-world graph applications.

\section{Conclusions and Outlook}
Based on the observations of intra-group accuracies and node embeddings, we find that nodes within a single real-world graph exhibit certain differences. It indicates that the weight-sharing scheme in GCN is not reliable. Thus we explore a spatial method with separated learning, NCGCN, which is totally guided by our proposed metric \nc. 
Our method addresses two common problems in existing spatial GNNs based on node separation: 1) the separation metrics are not appropriate; 2) training steps and separation strategies are complex and inefficient.

Our framework is simple yet effective enough to be a new start for spatial GNNs. The backbone can be replaced by more message passing models. The robustness and debiasing potential are also open to be explored.
Furthermore, \nc is an easy-to-compute metric that can effectively separate nodes for different optimization processes. This metric can also be further utilized in data mining, predictability, and interpretability.
The limitation of NCGCN is that it utilizes only second-order information following traditional GNNs. It seems sufficient on most real-world datasets except for long-distance dependent ones. Another problem is that high-\nc nodes are relatively rare on homophilous graphs, and we can use graph data augmentation~\cite{zhou2022data} for this issue in the future.

\bibliographystyle{IEEEtran}
\bibliography{Reference,IEEEabrv}

\begin{thebibliography}{10}
\providecommand{\url}[1]{#1}
\csname url@samestyle\endcsname
\providecommand{\newblock}{\relax}
\providecommand{\bibinfo}[2]{#2}
\providecommand{\BIBentrySTDinterwordspacing}{\spaceskip=0pt\relax}
\providecommand{\BIBentryALTinterwordstretchfactor}{4}
\providecommand{\BIBentryALTinterwordspacing}{\spaceskip=\fontdimen2\font plus
\BIBentryALTinterwordstretchfactor\fontdimen3\font minus \fontdimen4\font\relax}
\providecommand{\BIBforeignlanguage}[2]{{%
\expandafter\ifx\csname l@#1\endcsname\relax
\typeout{** WARNING: IEEEtran.bst: No hyphenation pattern has been}%
\typeout{** loaded for the language `#1'. Using the pattern for}%
\typeout{** the default language instead.}%
\else
\language=\csname l@#1\endcsname
\fi
#2}}
\providecommand{\BIBdecl}{\relax}
\BIBdecl

\bibitem{ma2021comprehensive}
X.~Ma, J.~Wu, S.~Xue, J.~Yang, C.~Zhou, Q.~Z. Sheng, H.~Xiong, and L.~Akoglu, ``A comprehensive survey on graph anomaly detection with deep learning,'' \emph{IEEE Trans. Knowl. Data Eng.}, vol.~35, no.~12, pp. 12\,012--12\,038, 2021.

\bibitem{luo2021future}
W.~Luo, W.~Liu, D.~Lian, and S.~Gao, ``Future frame prediction network for video anomaly detection,'' \emph{IEEE Trans. Pattern Anal. Mach. Intell.}, vol.~44, no.~11, pp. 7505--7520, 2021.

\bibitem{ahmed2021graph}
I.~Ahmed, T.~Galoppo, X.~Hu, and Y.~Ding, ``Graph regularized autoencoder and its application in unsupervised anomaly detection,'' \emph{IEEE Trans. Pattern Anal. Mach. Intell.}, vol.~44, no.~8, pp. 4110--4124, 2021.

\bibitem{cheng2020graph}
D.~Cheng, X.~Wang, Y.~Zhang, and L.~Zhang, ``Graph neural network for fraud detection via spatial-temporal attention,'' \emph{IEEE Trans. Knowl. Data Eng.}, vol.~34, no.~8, pp. 3800--3813, 2020.

\bibitem{zhou2022behavior}
J.~Zhou, C.~Hu, J.~Chi, J.~Wu, M.~Shen, and Q.~Xuan, ``Behavior-aware account de-anonymization on ethereum interaction graph,'' \emph{IEEE Trans. Inf. Forensics Security}, vol.~17, pp. 3433--3448, 2022.

\bibitem{10490264}
C.~Jin, J.~Zhou, J.~Jin, J.~Wu, and Q.~Xuan, ``Time-aware metapath feature augmentation for ponzi detection in ethereum,'' \emph{IEEE Trans. Netw. Sci. Eng.}, vol.~11, no.~4, pp. 3747--3758, 2024.

\bibitem{lyu2022knowledge}
Z.~Lyu, Y.~Wu, J.~Lai, M.~Yang, C.~Li, and W.~Zhou, ``Knowledge enhanced graph neural networks for explainable recommendation,'' \emph{IEEE Trans. Knowl. Data Eng.}, vol.~35, no.~5, pp. 4954--4968, 2022.

\bibitem{melton2023muxgnn}
J.~Melton and S.~Krishnan, ``muxgnn: Multiplex graph neural network for heterogeneous graphs,'' \emph{IEEE Trans. Pattern Anal. Mach. Intell.}, vol.~45, no.~9, pp. 11\,067--11\,078, 2023.

\bibitem{cai2021line}
L.~Cai, J.~Li, J.~Wang, and S.~Ji, ``Line graph neural networks for link prediction,'' \emph{IEEE Trans. Pattern Anal. Mach. Intell.}, vol.~44, no.~9, pp. 5103--5113, 2021.

\bibitem{yan2024federated}
B.~Yan, Y.~Cao, H.~Wang, W.~Yang, J.~Du, and C.~Shi, ``Federated heterogeneous graph neural network for privacy-preserving recommendation,'' in \emph{Proceedings of the ACM Web Conference}, 2024, pp. 3919--3929.

\bibitem{kipf2016semi}
T.~N. Kipf and M.~Welling, ``Semi-supervised classification with graph convolutional networks,'' in \emph{International Conference on Learning Representations}, 2022, pp. 1--14.

\bibitem{hamilton2017inductive}
W.~Hamilton, Z.~Ying, and J.~Leskovec, ``Inductive representation learning on large graphs,'' in \emph{Advances in Neural Information Processing Systems}, vol.~30, 2017, pp. 1--11.

\bibitem{velivckovic2017gat}
P.~Veli{\v{c}}kovi{\'c}, G.~Cucurull, A.~Casanova, A.~Romero, P.~Lio, and Y.~Bengio, ``Graph attention networks,'' in \emph{International Conference on Learning Representations}, 2018, pp. 1--12.

\bibitem{liu2022ud}
Y.~Liu, X.~Ao, F.~Feng, and Q.~He, ``Ud-gnn: Uncertainty-aware debiased training on semi-homophilous graphs,'' in \emph{Proceedings of the 28th ACM SIGKDD Conference on Knowledge Discovery and Data Mining}, 2022, pp. 1131--1140.

\bibitem{zheng2022survey}
X.~Zheng, Y.~Liu, S.~Pan, M.~Zhang, D.~Jin, and P.~S. Yu, ``Graph neural networks for graphs with heterophily: A survey,'' \emph{arXiv preprint arXiv:2202.07082}, 2022.

\bibitem{zheng2023node}
S.~Zheng, Z.~Zhu, Z.~Liu, Y.~Li, and Y.~Zhao, ``Node-oriented spectral filtering for graph neural networks,'' \emph{IEEE Trans. Pattern Anal. Mach. Intell.}, vol.~46, no.~1, pp. 388--402, 2024.

\bibitem{wang2021graph}
R.~Wang, S.~Mou, X.~Wang, W.~Xiao, Q.~Ju, C.~Shi, and X.~Xie, ``Graph structure estimation neural networks,'' in \emph{Proceedings of the ACM Web Conference}, 2021, pp. 342--353.

\bibitem{mao2023demystifying}
H.~Mao, Z.~Chen, W.~Jin, H.~Han, Y.~Ma, T.~Zhao, N.~Shah, and J.~Tang, ``Demystifying structural disparity in graph neural networks: Can one size fit all?'' in \emph{Advances in Neural Information Processing Systems}, vol.~36, 2023, pp. 37\,013--37\,067.

\bibitem{li2023restructuring}
S.~Li, D.~Kim, and Q.~Wang, ``Restructuring graph for higher homophily via adaptive spectral clustering,'' in \emph{Proceedings of the AAAI Conference on Artificial Intelligence}, vol.~37, no.~7, 2023, pp. 8622--8630.

\bibitem{chien2020adaptive}
E.~Chien, J.~Peng, P.~Li, and O.~Milenkovic, ``Adaptive universal generalized pagerank graph neural network,'' in \emph{International Conference on Learning Representations}, 2021, pp. 1--24.

\bibitem{he2021bernnet}
M.~He, Z.~Wei, H.~Xu \emph{et~al.}, ``Bernnet: Learning arbitrary graph spectral filters via bernstein approximation,'' in \emph{Advances in Neural Information Processing Systems}, vol.~34, 2021, pp. 14\,239--14\,251.

\bibitem{chen2020gcnii}
M.~Chen, Z.~Wei, Z.~Huang, B.~Ding, and Y.~Li, ``Simple and deep graph convolutional networks,'' in \emph{International conference on machine learning}.\hskip 1em plus 0.5em minus 0.4em\relax PMLR, 2020, pp. 1725--1735.

\bibitem{zhou2024pathmlp}
J.~Zhou, C.~Xie, S.~Gong, J.~Qian, S.~Yu, Q.~Xuan, and X.~Yang, ``Pathmlp: Smooth path towards high-order homophily,'' \emph{Neural Networks}, vol. 180, p. 106650, 2024.

\bibitem{luan2022acm}
S.~Luan, C.~Hua, Q.~Lu, J.~Zhu, M.~Zhao, S.~Zhang, X.-W. Chang, and D.~Precup, ``Revisiting heterophily for graph neural networks,'' in \emph{Advances in neural information processing systems}, vol.~35, 2022, pp. 1362--1375.

\bibitem{bo2021fagcn}
D.~Bo, X.~Wang, C.~Shi, and H.~Shen, ``Beyond low-frequency information in graph convolutional networks,'' in \emph{Proceedings of the AAAI conference on artificial intelligence}, vol.~35, no.~5, 2021, pp. 3950--3957.

\bibitem{wang2022hog}
T.~Wang, D.~Jin, R.~Wang, D.~He, and Y.~Huang, ``Powerful graph convolutional networks with adaptive propagation mechanism for homophily and heterophily,'' in \emph{Proceedings of the AAAI conference on artificial intelligence}, vol.~36, no.~4, 2022, pp. 4210--4218.

\bibitem{wuenergy}
Q.~Wu, Y.~Chen, C.~Yang, and J.~Yan, ``Energy-based out-of-distribution detection for graph neural networks,'' in \emph{International Conference on Learning Representations}, 2023, pp. 1--17.

\bibitem{li2022graphde}
Z.~Li, Q.~Wu, F.~Nie, and J.~Yan, ``Graphde: A generative framework for debiased learning and out-of-distribution detection on graphs,'' in \emph{Advances in Neural Information Processing Systems}, vol.~35, 2022, pp. 30\,277--30\,290.

\bibitem{bornmann2008citation}
L.~Bornmann and H.-D. Daniel, ``What do citation counts measure? a review of studies on citing behavior,'' \emph{Journal of documentation}, vol.~64, no.~1, pp. 45--80, 2008.

\bibitem{du2022gbk}
L.~Du, X.~Shi, Q.~Fu, X.~Ma, H.~Liu, S.~Han, and D.~Zhang, ``Gbk-gnn: Gated bi-kernel graph neural networks for modeling both homophily and heterophily,'' in \emph{Proceedings of the ACM Web Conference 2022}, 2022, pp. 1550--1558.

\bibitem{yan2022two}
Y.~Yan, M.~Hashemi, K.~Swersky, Y.~Yang, and D.~Koutra, ``Two sides of the same coin: Heterophily and oversmoothing in graph convolutional neural networks,'' in \emph{2022 IEEE International Conference on Data Mining (ICDM)}.\hskip 1em plus 0.5em minus 0.4em\relax IEEE, 2022, pp. 1287--1292.

\bibitem{pei2019geom}
H.~Pei, B.~Wei, K.~C.-C. Chang, Y.~Lei, and B.~Yang, ``Geom-gcn: Geometric graph convolutional networks,'' in \emph{International Conference on Learning Representations}, 2019, pp. 1--12.

\bibitem{ma2021necessity}
Y.~Ma, X.~Liu, N.~Shah, and J.~Tang, ``Is homophily a necessity for graph neural networks?'' in \emph{International Conference on Learning Representations}, 2021, pp. 1--28.

\bibitem{zhu2020h2gcn}
J.~Zhu, Y.~Yan, L.~Zhao, M.~Heimann, L.~Akoglu, and D.~Koutra, ``Beyond homophily in graph neural networks: Current limitations and effective designs,'' in \emph{Advances in Neural Information Processing Systems}, vol.~33, 2020, pp. 7793--7804.

\bibitem{zhang2022degradation}
W.~Zhang, Z.~Sheng, Z.~Yin, Y.~Jiang, Y.~Xia, J.~Gao, Z.~Yang, and B.~Cui, ``Model degradation hinders deep graph neural networks,'' in \emph{Proceedings of the 28th ACM SIGKDD conference on knowledge discovery and data mining}, 2022, pp. 2493--2503.

\bibitem{oono2019graph}
K.~Oono and T.~Suzuki, ``Graph neural networks exponentially lose expressive power for node classification,'' in \emph{International Conference on Learning Representations}, 2019, pp. 1--37.

\bibitem{li2018deeper}
Q.~Li, Z.~Han, and X.-M. Wu, ``Deeper insights into graph convolutional networks for semi-supervised learning,'' in \emph{Proceedings of the AAAI conference on artificial intelligence}, vol.~32, no.~1, 2018.

\bibitem{abu2019mixhop}
S.~Abu-El-Haija, B.~Perozzi, A.~Kapoor, N.~Alipourfard, K.~Lerman, H.~Harutyunyan, G.~Ver~Steeg, and A.~Galstyan, ``Mixhop: Higher-order graph convolutional architectures via sparsified neighborhood mixing,'' in \emph{International Conference on Machine Learning}.\hskip 1em plus 0.5em minus 0.4em\relax PMLR, 2019, pp. 21--29.

\bibitem{luan2019snowball}
S.~Luan, M.~Zhao, X.-W. Chang, and D.~Precup, ``Break the ceiling: Stronger multi-scale deep graph convolutional networks,'' in \emph{Advances in Neural Information Processing Systems}, vol.~32, 2019, pp. 1--11.

\bibitem{zhang2019survey}
S.~Zhang, H.~Tong, J.~Xu, and R.~Maciejewski, ``Graph convolutional networks: a comprehensive review,'' \emph{Computational Social Networks}, vol.~6, no.~1, pp. 1--23, 2019.

\bibitem{suresh2021wrgat}
S.~Suresh, V.~Budde, J.~Neville, P.~Li, and J.~Ma, ``Breaking the limit of graph neural networks by improving the assortativity of graphs with local mixing patterns,'' in \emph{Proceedings of the 27th ACM SIGKDD conference on knowledge discovery \& data mining}, 2021, pp. 1541--1551.

\bibitem{zou2023similarity}
M.~Zou, Z.~Gan, R.~Cao, C.~Guan, and S.~Leng, ``Similarity-navigated graph neural networks for node classification,'' \emph{Inform. Sciences}, vol. 633, pp. 41--69, 2023.

\bibitem{cavallo2023NCS}
A.~Cavallo, G.~Claas, R.~Michele, L.~Giulio, L.~Vassio \emph{et~al.}, ``2-hop neighbor class similarity (2ncs): A graph structural metric indicative of graph neural network performance,'' in \emph{3rd Workshop on Graphs and more Complex structures for Learning and Reasoning (GCLR) at AAAI 2023}, 2023.

\bibitem{cagnn}
J.~Chen, S.~Chen, J.~Gao, Z.~Huang, J.~Zhang, and J.~Pu, ``Exploiting neighbor effect: Conv-agnostic gnn framework for graphs with heterophily,'' \emph{IEEE Trans. Neural Netw. Learn. Syst.}, vol.~35, no.~10, pp. 13\,383--13\,396, 2023.

\bibitem{wang2021bag}
Y.~Wang, ``Bag of tricks of semi-supervised classification with graph neural networks,'' \emph{arXiv preprint arXiv:2103.13355}, 2021.

\bibitem{xiaoliu}
Y.~Gao, X.~Wang, X.~He, Z.~Liu, H.~Feng, and Y.~Zhang, ``Addressing heterophily in graph anomaly detection: A perspective of graph spectrum,'' in \emph{Proceedings of the ACM Web Conference 2023}, 2023, pp. 1528--1538.

\bibitem{zhu2021cpgnn}
J.~Zhu, R.~A. Rossi, A.~Rao, T.~Mai, N.~Lipka, N.~K. Ahmed, and D.~Koutra, ``Graph neural networks with heterophily,'' in \emph{Proceedings of the AAAI conference on artificial intelligence}, vol.~35, no.~12, 2021, pp. 11\,168--11\,176.

\bibitem{lwgcn}
E.~Dai, S.~Zhou, Z.~Guo, and S.~Wang, ``Label-wise graph convolutional network for heterophilic graphs,'' in \emph{Learning on Graphs Conference}.\hskip 1em plus 0.5em minus 0.4em\relax PMLR, 2022, pp. 26--1.

\bibitem{shchur2018pitfalls}
O.~Shchur, M.~Mumme, A.~Bojchevski, and S.~G{\"u}nnemann, ``Pitfalls of graph neural network evaluation,'' \emph{arXiv preprint arXiv:1811.05868}, 2018.

\bibitem{fano1961transmission}
R.~M. Fano and D.~Hawkins, ``Transmission of information: A statistical theory of communications,'' \emph{American Journal of Physics}, vol.~29, no.~11, pp. 793--794, 1961.

\bibitem{klicpera2019diffusion}
J.~Klicpera, S.~Wei{\ss}enberger, and S.~G{\"u}nnemann, ``Diffusion improves graph learning,'' in \emph{Advances in Neural Information Processing Systems}, 2019, pp. 13\,366--13\,378.

\bibitem{lim2021linkx}
D.~Lim, F.~Hohne, X.~Li, S.~L. Huang, V.~Gupta, O.~Bhalerao, and S.~N. Lim, ``Large scale learning on non-homophilous graphs: New benchmarks and strong simple methods,'' in \emph{Advances in Neural Information Processing Systems}, vol.~34, 2021, pp. 20\,887--20\,902.

\bibitem{critical}
O.~Platonov, D.~Kuznedelev, M.~Diskin, A.~Babenko, and L.~Prokhorenkova, ``A critical look at the evaluation of gnns under heterophily: Are we really making progress?'' in \emph{International Conference on Learning Representations}, 2023, pp. 1--15.

\bibitem{li2022glognn}
X.~Li, R.~Zhu, Y.~Cheng, C.~Shan, S.~Luo, D.~Li, and W.~Qian, ``Finding global homophily in graph neural networks when meeting heterophily,'' in \emph{International Conference on Machine Learning}.\hskip 1em plus 0.5em minus 0.4em\relax PMLR, 2022, pp. 13\,242--13\,256.

\bibitem{kingma2014adam}
D.~P. Kingma and J.~Ba, ``Adam: A method for stochastic optimization,'' \emph{arXiv preprint arXiv:1412.6980}, 2014.

\bibitem{wang2020microsoft}
K.~Wang, Z.~Shen, C.~Huang, C.-H. Wu, Y.~Dong, and A.~Kanakia, ``Microsoft academic graph: When experts are not enough,'' \emph{Quantitative Science Studies}, vol.~1, no.~1, pp. 396--413, 2020.

\bibitem{Zeng2020GraphSAINT}
H.~Zeng, H.~Zhou, A.~Srivastava, R.~Kannan, and V.~Prasanna, ``Graphsaint: Graph sampling based inductive learning method,'' in \emph{International Conference on Learning Representations}, 2020, pp. 1--19.

\bibitem{zhou2022data}
J.~Zhou, C.~Xie, S.~Gong, Z.~Wen, X.~Zhao, Q.~Xuan, and X.~Yang, ``Data augmentation on graphs: a technical survey,'' \emph{arXiv preprint arXiv:2212.09970}, 2022.

\bibitem{cora}
P.~Sen, G.~Namata, M.~Bilgic, L.~Getoor, B.~Galligher, and T.~Eliassi-Rad, ``Collective classification in network data,'' \emph{AI magazine}, vol.~29, no.~3, pp. 93--93, 2008.

\bibitem{yang2016revisiting}
Z.~Yang, W.~Cohen, and R.~Salakhudinov, ``Revisiting semi-supervised learning with graph embeddings,'' in \emph{International conference on machine learning}.\hskip 1em plus 0.5em minus 0.4em\relax PMLR, 2016, pp. 40--48.

\bibitem{coraf}
A.~Bojchevski and S.~G{\"u}nnemann, ``Deep gaussian embedding of graphs: Unsupervised inductive learning via ranking,'' in \emph{International Conference on Learning Representations}, 2018, pp. 1--13.

\bibitem{photo}
J.~McAuley, C.~Targett, Q.~Shi, and A.~Van Den~Hengel, ``Image-based recommendations on styles and substitutes,'' in \emph{Proceedings of the 38th international ACM SIGIR conference on research and development in information retrieval}, 2015, pp. 43--52.

\bibitem{nni2021}
\BIBentryALTinterwordspacing
{Microsoft}, ``{Neural Network Intelligence},'' 2021. [Online]. Available: \url{https://github.com/microsoft/nni}
\BIBentrySTDinterwordspacing

\bibitem{bergstra2011algorithms}
J.~Bergstra, R.~Bardenet, Y.~Bengio, and B.~K{\'e}gl, ``Algorithms for hyper-parameter optimization,'' in \emph{NeurIPS}, vol.~24, 2011, pp. 1--9.

\bibitem{APPNP}
J.~Gasteiger, A.~Bojchevski, and S.~G{\"u}nnemann, ``Predict then propagate: Graph neural networks meet personalized pagerank,'' in \emph{ICLR}, 2019.

\bibitem{luo2021colink}
Y.~Luo, A.~Chen, K.~Yan, and L.~Tian, ``Distilling self-knowledge from contrastive links to classify graph nodes without passing messages,'' \emph{arXiv preprint arXiv:2106.08541}, 2021.

\bibitem{3ference}
Y.~Luo, G.~Luo, K.~Yan, and A.~Chen, ``Inferring from references with differences for semi-supervised node classification on graphs,'' \emph{Mathematics}, vol.~10, no.~8, p. 1262, 2022.

\bibitem{shirzad2023exphormer}
H.~Shirzad, A.~Velingker, B.~Venkatachalam, D.~J. Sutherland, and A.~K. Sinop, ``Exphormer: Sparse transformers for graphs,'' in \emph{International Conference on Machine Learning}.\hskip 1em plus 0.5em minus 0.4em\relax PMLR, 2023, pp. 31\,613--31\,632.

\bibitem{Fav}
Y.~Guo and Z.~Wei, ``Graph neural networks with learnable and optimal polynomial bases,'' in \emph{International Conference on Machine Learning}.\hskip 1em plus 0.5em minus 0.4em\relax PMLR, 2023, pp. 12\,077--12\,097.

\end{thebibliography}

% \newpage

\vspace{-40pt}
\begin{IEEEbiography}[{\includegraphics[width=1in,height=1.25in,clip,keepaspectratio]{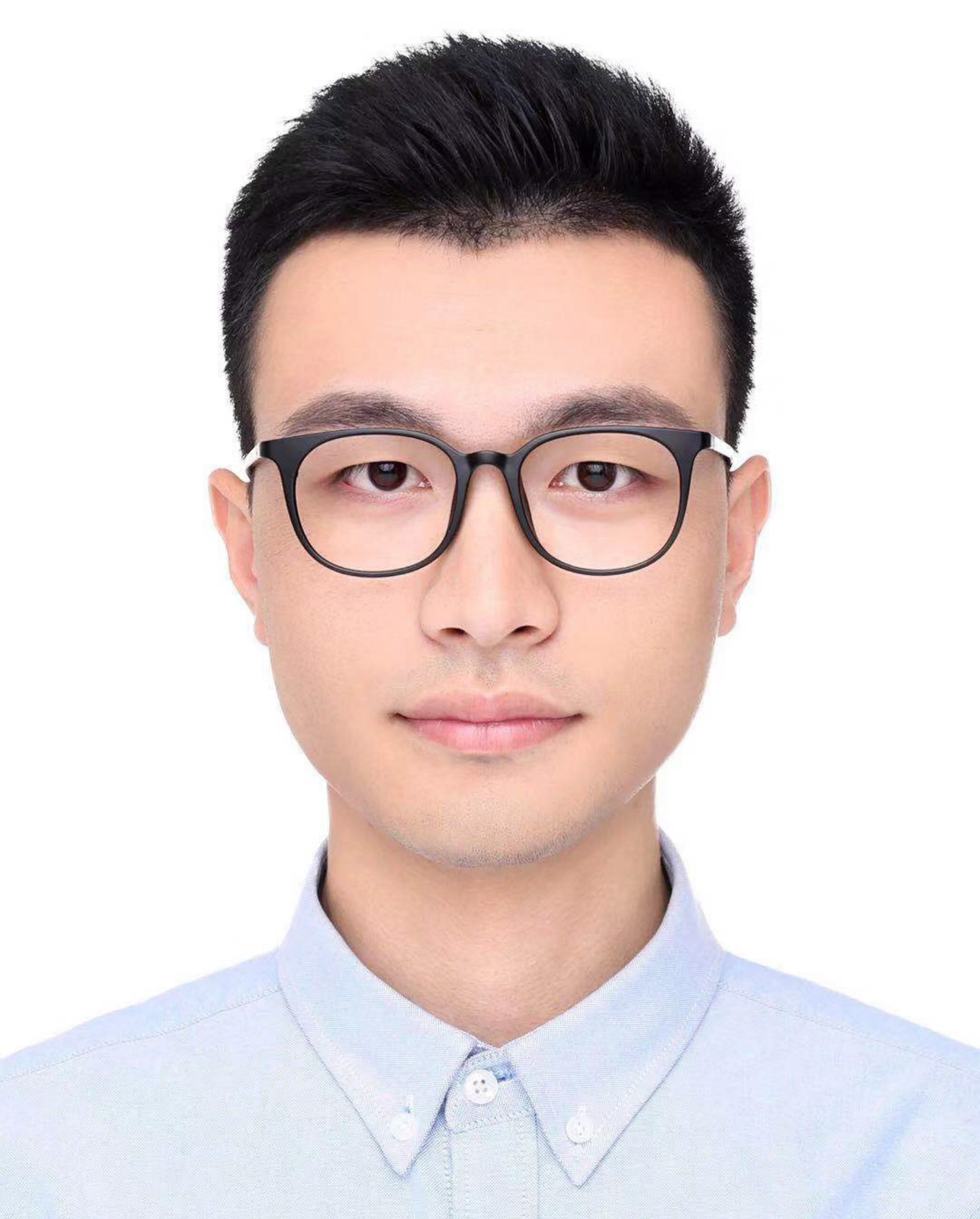}}]{Jiajun Zhou}
	received the Ph.D degree in control theory and engineering from Zhejiang University of Technology, Hangzhou, China, in 2023. He is currently a Postdoctoral Research Fellow with the Institute of Cyberspace Security, Zhejiang University of Technology. His current research interests include graph data mining, cyberspace security and data management.
\end{IEEEbiography}
\vspace{-40pt}

\begin{IEEEbiography}[{\includegraphics[width=1in,height=1.25in,clip,keepaspectratio]{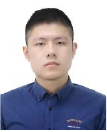}}]{Shengbo Gong}
	received the BS degree from Shanghai Jiaotong University, Shanghai, China, in 2020. He is currently pursuing the MS degree in control engineering at Zhejiang University of Technology, Hangzhou, China. His current research interests include graph data mining and graph neural network, especially for heterophily problem in graph.
\end{IEEEbiography}
\vspace{-40pt}

\begin{IEEEbiography}[{\includegraphics[width=1in,height=1.25in,clip,keepaspectratio]{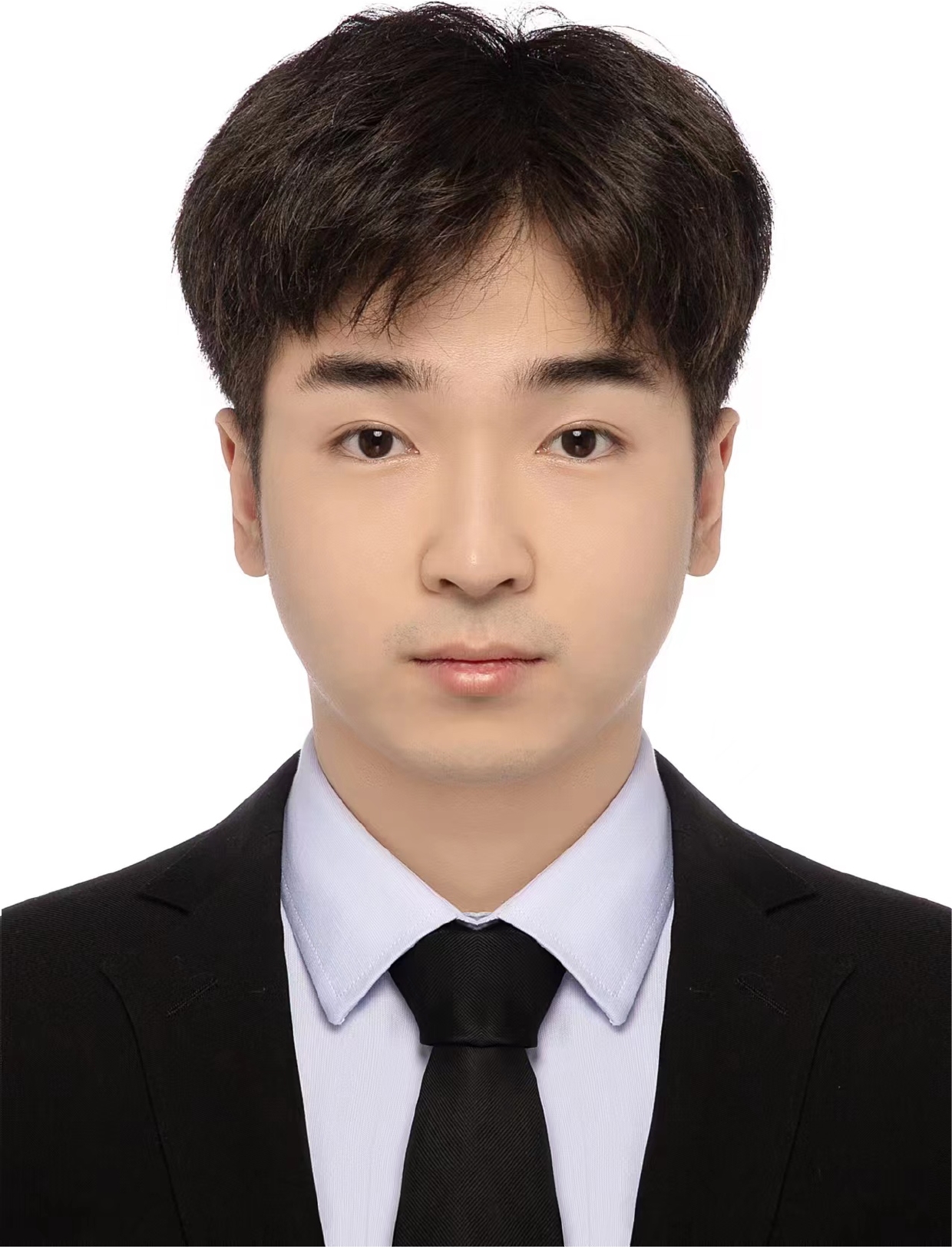}}]{Xuanze Chen}
	received the BS degree from Wenzhou University, Wenzhou, Zhejiang, China, in 2023. He is currently pursuing a Master's degree at the Institute of Cyberspace Security, Zhejiang University of Technology. His current research interests include graph neural networks and Graph Transformers.
\end{IEEEbiography}
\vspace{-40pt}

\begin{IEEEbiography}[{\includegraphics[width=1in,height=1.25in,clip,keepaspectratio]{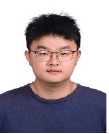}}]{Chenxuan Xie}
	received the BS degree in automation from Nanchang University, Jiangxi, China, in 2021. He is currently pursuing the MS degree in control engineering at Zhejiang University of Technology, Hangzhou, China. His current research interests include graph data mining and graph neural network, especially for graph heterophily problem and graph transformer.
\end{IEEEbiography}
\vspace{-40pt}

\begin{IEEEbiography}[{\includegraphics[width=1in,height=1.25in,clip,keepaspectratio]{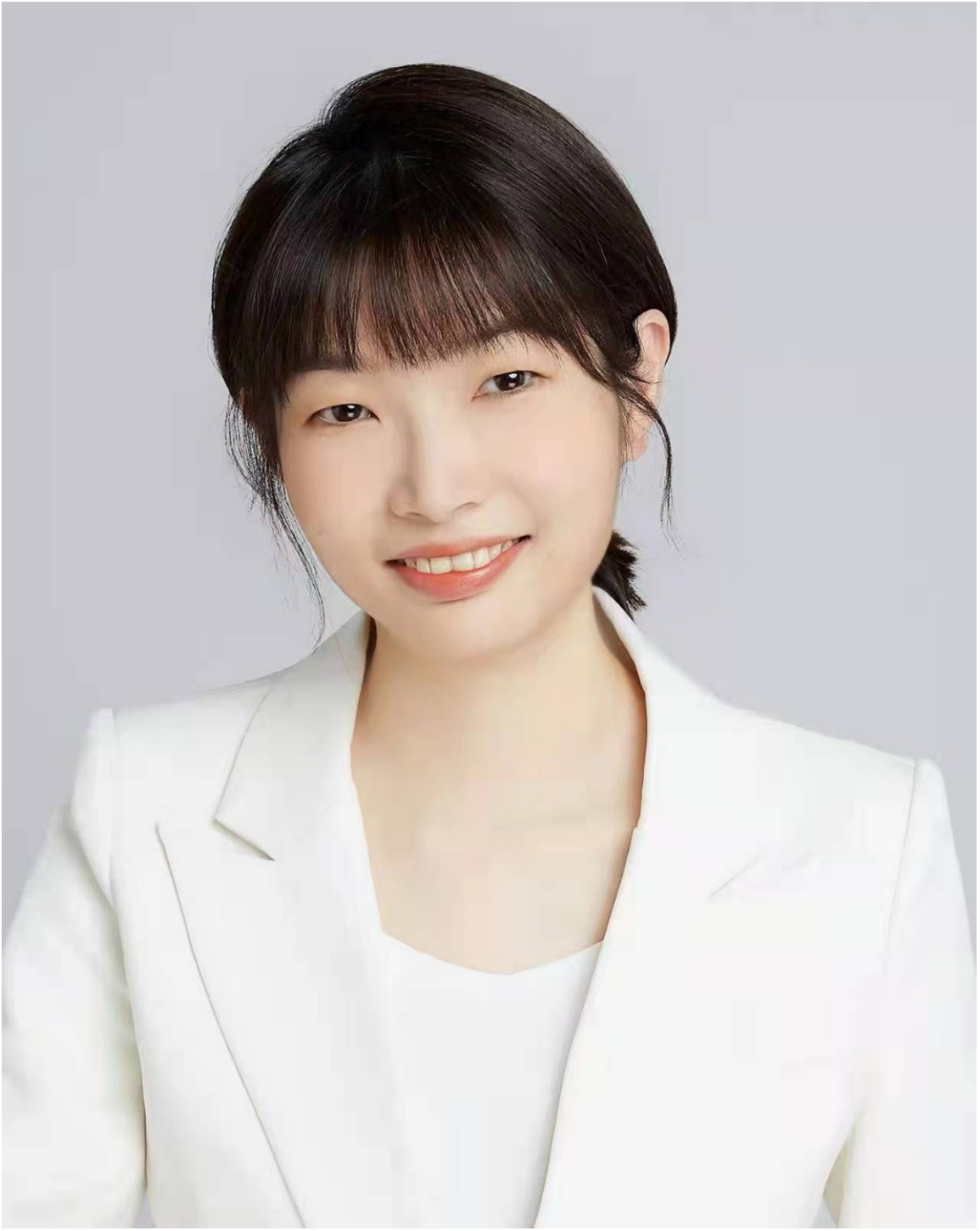}}]{Shanqing Yu}
	received the M.S. degree from the School of Computer Engineering and Science, Shanghai University, China, in 2008 and received the M.S. degree from the Graduate School of Information, Production and Systems, Waseda University, Japan, in 2008, and the Ph.D. degree, in 2011, respectively. She is currently a Lecturer at the Institute of Cyberspace Security and the College of Information Engineering, Zhejiang University of Technology, Hangzhou, China. Her research interests cover intelligent computation and data mining.
\end{IEEEbiography}
\vspace{-40pt}

\begin{IEEEbiography}[{\includegraphics[width=1in,height=1.25in,clip,keepaspectratio]{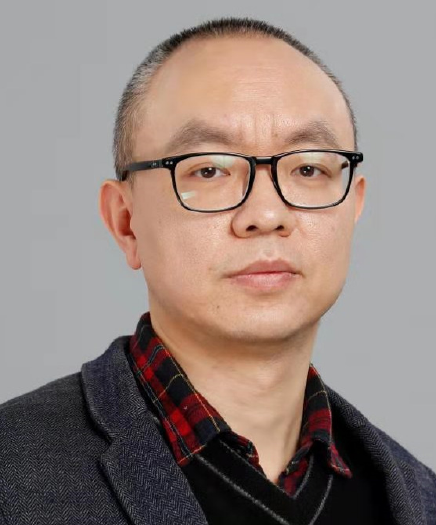}}]{Qi Xuan}(M'18) 
  received the BS and PhD degrees in control theory and engineering from Zhejiang University, Hangzhou, China, in 2003 and 2008, respectively. He was a Post-Doctoral Researcher with the Department of Information Science and Electronic Engineering, Zhejiang University, from 2008 to 2010, respectively, and a Research Assistant with the Department of Electronic Engineering, City University of Hong Kong, Hong Kong, in 2010 and 2017. From 2012 to 2014, he was a Post-Doctoral Fellow with the Department of Computer Science, University of California at Davis, CA, USA. He is a senior member of the IEEE and is currently a Professor with the Institute of Cyberspace Security, College of Information Engineering, Zhejiang University of Technology, Hangzhou, China. His current research interests include network science, graph data mining, cyberspace security, machine learning, and computer vision.
\end{IEEEbiography}
\vspace{-40pt}

\begin{IEEEbiography}[{\includegraphics[width=1in,height=1.25in,clip,keepaspectratio]{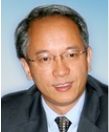}}]{Xiaoniu Yang}
    is currently a Chief Scientist with the Science and Technology on Communication
    Information Security Control Laboratory, Jiaxing, China. He is also an Academician of the Chinese Academy of Engineering and a fellow of the Chinese
    Institute of Electronics. He has published the first software radio book \emph{Software Radio Principles and Applications} [China: Publishing House of Electronics Industry, X. Yang, C. Lou, and J. Xu, 2001 (in Chinese)]. His current research interests are in software-defined satellites, big data for radio signals, and deep learning-based signal processing.
  \end{IEEEbiography}

% \newpage
% ~
\newpage

% \newpage
% \mbox{}
% \newpage

\appendix

\subsection{{\textbf{More Proof}}}\label{app:proof}
\begin{proposition}
    {\emph{The conditional entropy, which reflects classification uncertainty, is positively correlated with \nc within its effective range of $[0, 1]$, allowing the classification error rate to be theoretically bounded using \nc.}}
\end{proposition}
\begin{proof}
    {Directly analyzing the correlation between \nc and conditional entropy $H(y_i| Y_{\mathcal{N}_i^k})$ is mathematically complex. We can simplify the proof by leveraging the transitivity of correlation. Specifically, we first treat $p_{\max}$ as a function of \nc (Eq.~(\ref{eq: pmax nc})) and conditional entropy $H(y_i| Y_{\mathcal{N}_i^k})$ as a function of $p_{\max}$ (Eq.~(\ref{eq: tiaojian-pmax})).}

    {\textbf{Step 1: analyze the correlation between $p_{\max}$ and \nc.} Consider the derivative of $p_{\max}$ with respect to \nc:
    \begin{equation}
        \frac{d p_{\max }}{d \textit{NC}}=-\log |\mathcal{C}| \cdot 2^{-\textit{NC} \cdot \log |\mathcal{C}|} \cdot \ln 2<0
    \end{equation}
    This shows that $p_{\max}$ decreases as \nc increases, confirming the negative correlation between $p_{\max}$ and \nc. And $p_{\max}\in[\frac{1}{|\mathcal{C}|}, 1]$.}

    {\textbf{Step 2: analyze the correlation between $H(y_i| Y_{\mathcal{N}_i^k})$ and $p_{\max}$.} Consider the derivative of $H(y_i| Y_{\mathcal{N}_i^k})$ with respect to $p_{\max}$:
    \begin{equation}
        \begin{split}
            \frac{d H\left(y_i \mid Y_{\mathcal{N}_i^k}\right)}{d p_{\max} }=\log _2\left(\frac{1-p_{\max }}{|\mathcal{C}|-1} \cdot \frac{1}{p_{\max }}\right) \\
            =\log _2\left(\left(\frac{1}{p_{\max }}-1\right) \cdot \frac{1}{|\mathcal{C}|-1}\right) \leq 0
        \end{split}
    \end{equation}
    It is evident that this derivative is negatively correlated with $p_{\max}$, and its maximum value occurs when $p_{\max}=\frac{1}{|\mathcal{C}|}$, i.e., $\frac{d H(y_i | Y_{\mathcal{N}_i^k})}{d p_{\max} }|_{p_{\max}=\frac{1}{|\mathcal{C}|}}=0$ and $\frac{d H(y_i \mid Y_{\mathcal{N}_i^k})}{d p_{\max} }\leq 0$. This shows that $H(y_i | Y_{\mathcal{N}_i^k})$ decreases as $p_{\max}$ increases, confirming the negative correlation between $H(y_i | Y_{\mathcal{N}_i^k})$ and $p_{\max}$.}

    {\textbf{Step 3: conclude the correlation between $H(y_i | Y_{\mathcal{N}_i^k})$ and \nc.} From the above steps, we can conclude that $p_{\max}$ and \nc are negatively correlated, and $H(y_i | Y_{\mathcal{N}_i^k})$ and $p_{\max}$ are negatively correlated. By the transitive property of correlation, $H(y_i | Y_{\mathcal{N}_i^k})$ and \nc are positively correlated. As \nc increases, $H(y_i | Y_{\mathcal{N}_i^k})$ also increases.}

    {\textbf{Step 4: analyze the correlation between the lower bound function and \nc.} The lower bound function can be defined as follows:
    \begin{equation}
        \frac{H(y_i\mid Y_{\mathcal{N}_i^k})-1}{\log |\mathcal{C}|} = \frac{f(\textit{NC})-1}{\log |\mathcal{C}|}
    \end{equation}
    Its monotonicity is consistent with that of conditional entropy, meaning that as \nc increases, the lower bound of the classification error rate also rises. Therefore, the classification error rate can be theoretically bounded using \nc.}
\end{proof}

\begin{figure}[!ht]
    \centering
    \includegraphics[width = \linewidth]{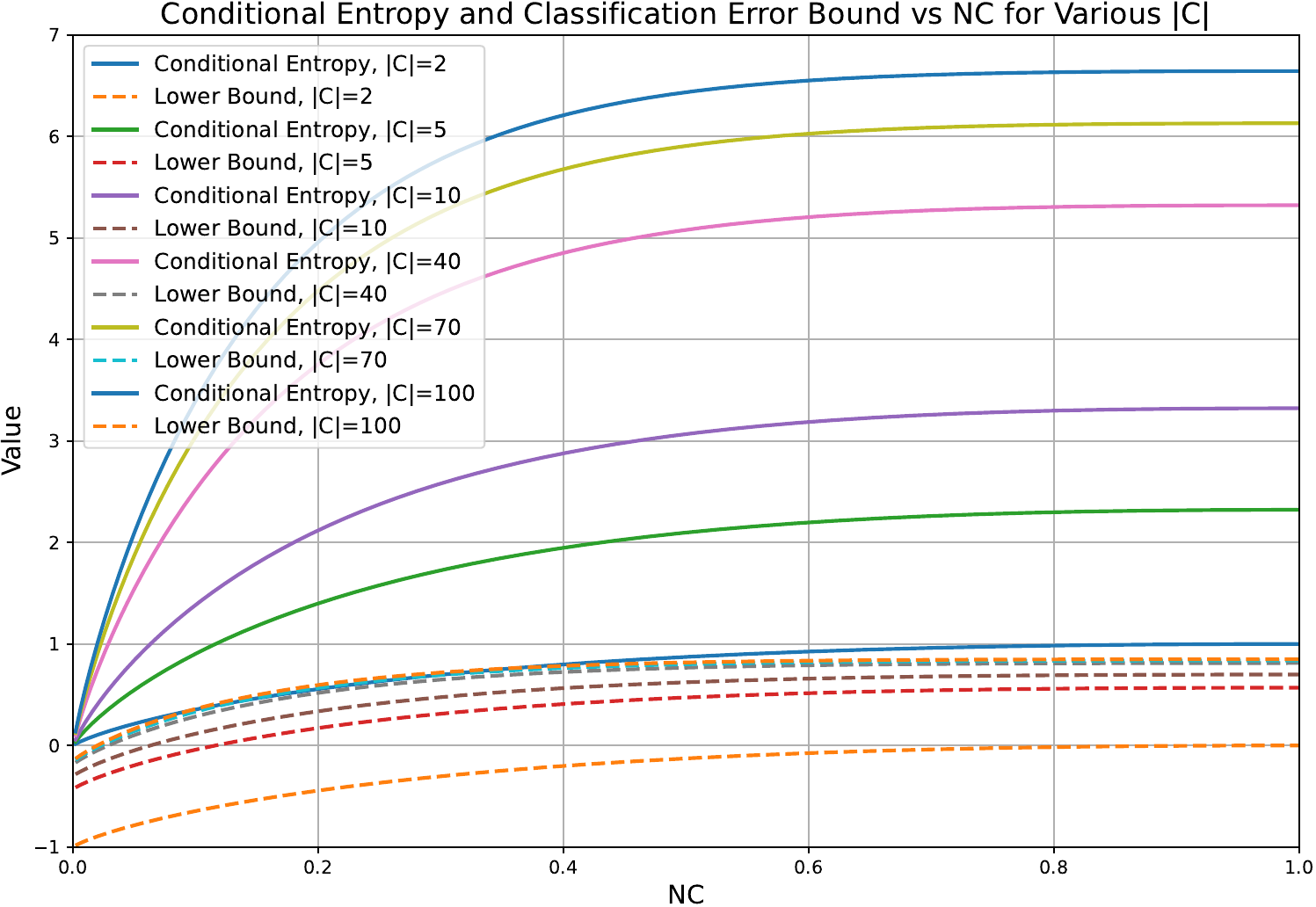}
    \caption{{The curve of conditional entropy and classification error bound versus \nc under different numbers of label categories.}}
    \label{fig: functon}
\end{figure}

\subsection{\textbf{Dataset Details}}\label{app:datasets}
\subsubsection{\textbf{Dataset Descriptions}}
To facilitate a better understanding of the dataset selection, some additional dataset descriptions are provided as follows.

\begin{itemize}%[leftmargin=10pt]
    \item \textbf{Citation networks.} 
    \texttt{Cora}, \texttt{Citeseer}, \texttt{Pubmed}~\cite{cora,yang2016revisiting} and \texttt{Cora Full}~\cite{coraf} are citation networks where nodes denote papers, edges denote their citation or cited relationships and node features are constructed based on keywords of papers. The first three graphs serve as benchmarks in almost all GNN-related papers. For small datasets like \texttt{Cora} and \texttt{Citeseer}, the highest results in public rankings may have been over-fitting, while larger ones like \texttt{Pubmed} and \texttt{Cora Full}, may not have been fully explored. In addition, \texttt{Pubmed} and \texttt{Cora Full} also show lower node homophily than the first two, indicating that they contain more complex patterns.
    \item \textbf{Amazon co-purchase networks.} 
    \texttt{Computers} and \texttt{Photo}~\cite{photo,shchur2018pitfalls} are segments of Amazon co-purchase network, in which node features are bag-of-words codes for product reviews, edges denote frequent co-purchases between goods, and labels denote product categories.
    \item \textbf{Co-authorship networks.} 
    \texttt{Coauthor CS} and \texttt{Coauthor Physics} are co-authorship networks based on the Microsoft Academic Network from the KDD Cup 2016 challenge~\cite{shchur2018pitfalls,wang2020microsoft}. 
    The nodes of two researchers will be connected if they have collaborated on any research paper. The node features represent the bag-of-words representations of the papers authored by the researchers, while the node label indicates their respective research fields.
    \item \textbf{Actor co-occurrence network.} 
    The \texttt{Actor} dataset is the actor-only induced subgraph of the film-director-actor-writer network, providing insights into the co-occurrence relationships among actors. The node features encompass keywords extracted from the actors' Wikipedia pages.
    \item \textbf{Wikipedia networks.} 
    The \texttt{Squirrel} and \texttt{Chameleon} datasets are sourced from Wikipedia page-page network, where nodes represent articles from the English Wikipedia (December 2018), and edges indicate mutual links between them. Node features consist of Bag-of-words representing specific nouns in the articles. Nodes are labeled by Pei et al.~\cite{pei2019geom} based on average monthly traffic. \emph{However, a recent study~\cite{critical} highlights that these two datasets contain a significant number of structurally duplicated nodes.}
    We follow this work and use \texttt{Squirrel Filtered} and \texttt{Chameleon Filtered} in their publicly available code as benchmarks.
    % \footnote{\href{https://github.com/yandex-research/heterophilous-graphs/tree/main/data}{https://github.com/yandex-research/heterophilous-graphs/tree/main/data}} 
    \item \textbf{Social networks.} 
    \texttt{Penn94}~\cite{lim2021linkx} is a subgraph extracted from Facebook, specifically focusing on students as nodes. These nodes are characterized by various features such as their major, second major/minor, dorm/house, year, and high school background. The nodes are labeled based on gender. Despite its significant heterophily and substantial size, this dataset has been overlooked by several studies addressing the heterophily problem.
\end{itemize}

\begin{table*}[!htb]
    \renewcommand\arraystretch{1.2}
    \centering
    \caption{Duplication detection for nodes of same labels. Boldface letters are used to mark benchmarks with data leakage issue.}
    \label{tab:duplication}
    \resizebox{\linewidth}{!}{
    \begin{tabular}{ccccccccccccc} 
    \hline\hline
    Duplication rate (\%)      & \textbf{Pubmed} & \textbf{Photo} & \textbf{Computers} & \begin{tabular}[c]{@{}c@{}}\textbf{Coauthor} \\\textbf{CS}\end{tabular} & \begin{tabular}[c]{@{}c@{}}\textbf{Coauthor} \\\textbf{Physics}\end{tabular} & \textbf{Cora Full} & \textbf{Actor} & \textbf{Penn94} & \begin{tabular}[c]{@{}c@{}}\textbf{Chameleon}\\\textbf{Filtered}\end{tabular} & \begin{tabular}[c]{@{}c@{}}\textbf{Squirrel}\\\textbf{Filtered}\end{tabular} & \textbf{Chameleon} & \textbf{Squirrel}  \\ 
    \hline
    neighbors' index + labels & 29.46           & 1.58           & 2.18               & 0.65                                                                    & 0.22                                                                         & 3.96               & 14.95          & 0.04            & 1.85                                                                          & 0.09                                                                         & \textbf{46.03}     & \textbf{36.28}     \\
    raw features + labels     & 0.02            & 0.08           & 0.04               & 0.23                                                                    & 1.02                                                                         & 1.32               & 14.11          & 7.13            & 10.76                                                                         & 5.80                                                                         & 13.31              & 3.65               \\
    \hline\hline
    \end{tabular}}
\end{table*}

\subsubsection{\textbf{Duplication Detection}}
% A recent work~\cite{critical} points out that the two datasets (\texttt{Squirrel} and \texttt{Chameleon}) contain too many duplicated nodes structurally (almost 50\%), which leads to data leakage in methods using adjacent matrices, such as ACM++~\cite{luan2022acm} and LINKX~\cite{lim2021linkx}. After filtering duplicated nodes, the performance of these methods degrades significantly. Aware of the importance of this issue, we detect the duplication rate for all datasets. The results are shown in Table~\ref{tab:duplication}.
A recent study~\cite{critical} highlights that the two datasets (\texttt{Squirrel} and \texttt{Chameleon}) exhibit a significant structural duplication of nodes (almost 50\%), resulting in data leakage issues for methods utilizing adjacency matrices, such as ACM++~\cite{luan2022acm} and LINKX~\cite{lim2021linkx}. The performance of these methods experiences a substantial degradation after eliminating the duplicated nodes. Recognizing the criticality of this concern, we assess the duplication rate across all datasets. The corresponding results are presented in Table~\ref{tab:duplication}.

\subsubsection{\textbf{Criteria for Dataset Selection}}
Inspired by~\cite{critical}, we carefully re-evaluate datasets widely used in the field of heterophily problem. To circumvent the drawbacks of under-sampling and data leakage while preserving the characteristics of an authentic dataset, our selection criteria are as follows: 1) the number of nodes should exceed 100 times the number of classes; 
2) the duplication rate should not surpass 30\%. 
Based on criterion 1, we exclude small datasets like \texttt{Texas}. 
Based on criterion 2, we eliminate \texttt{Chameleon} and \texttt{Squirrel}. 
Finally, we select 10 datasets with varying proportions of heterophilous nodes to emphasize the node-level generality of our framework. To facilitate comparison with previous studies, these datasets are still categorized into homophilous and heterophilous groups based on their average \textit{NH} values.

\subsubsection{\textbf{Dataset Sources and Split Settings}}
We use the datasets from GPRGNN~\cite{chien2020adaptive}, except \texttt{Chameleon} and \texttt{Squirrel}, which are from~\cite{critical}. For the training/validation/testing split, we follow GPRGNN due to its extensive validation and availability as an open-source implementation. It contains a class-balanced training set and fully shuffled validation and testing sets. 
The ACM-GCN method~\cite{luan2022acm} adopts some datasets with a different number of edges compared to other methods, necessitating the re-execution of the ACM-GCN source code on the dataset used by GPRGNN.
Exploring subtle discrepancies within the datasets is beyond the scope of this paper.

\subsection{\textbf{Baseline Details and More Comparisons}}\label{app:baselines}
\subsubsection{\textbf{Baseline Methods}}
To facilitate a better understanding of the baseline selection, some additional descriptions of baseline methods are provided as follows
\begin{itemize}
    \item \textbf{Two-layer MLP} is a neural network that only focuses on learning node representations from raw features, without propagation and aggregation rules.
    \item \textbf{GCN} is a classical message passing network that aggregates information from neighbors on average.
    \item \textbf{GAT} is a message passing network that aggregates information from neighbors by multi-head attention.
    \item \textbf{SAGE} is a message passing network that concatenates the transformed features of center nodes with the average features of their neighboring nodes.
    \item \textbf{H2GCN} combines ego-, low- and high-order features, and finally concatenates the intermediate representations before readout.
    % \item \textbf{GCNII} is a graph neural network using two effective tricks on the basis of GCN: initial residual and identity mapping.
    % \item \textbf{BernNet} is a graph neural network that uses Bernstein polynomial approximation to design frequency filters.
    \item \textbf{GPRGNN} adaptively optimizes the Generalized PageRank weights to control the contribution of the propagation process at each layer.
    \item \textbf{FAGCN} is a frequency-adaptive graph neural network that can adaptively aggregate low-frequency and high-frequency information.
    \item \textbf{ACM-GCN} adaptively mixes low-pass, high-pass, and identity channels with node-wise weights to extract more neighborhood information.
    \item \textbf{GBKSage} mixes homophilous and heterophilous transformation matrices by appropriate weights from a gating mechanism. 
    \item \textbf{HOG} utilizes topology and attribute information of graphs to learn the homophily degree between each node pair, thereby enabling modifications for propagation rules.
    \item \textbf{WRGAT} lets structurally similar nodes pass messages to each other by building a multi-relational graph on top of the original graph.
    \item \textbf{CPGNN} uses label transition matrices to fix transition probabilities between two classes of nodes. It is cold-started by pre-training and pseudo labels.
    \item \textbf{GloGNN} independently embeds and integrates adjacency information and node features using MLPs. Subsequently, it computes the similarity between each pair of nodes to derive a coefficient matrix, which can be seen as a rewired and soft adjacency matrix.
    \item \textbf{SNGNN} uses node similarity matrix coupled with mean aggregation in the neighborhood aggregation process.
    \item \textbf{CAGNN} decouples the node features into discriminative features for downstream tasks and aggregation features, followed by employing a shared mixer module to adaptively evaluate the impact of neighboring nodes.
\end{itemize}

\subsubsection{\textbf{Criteria for Baseline Selection}}\label{app:standard-baseline}
Firstly, we compare NCGCN with spatial methods that optimize message passing through rewiring, gating mechanisms, label propagation, and other techniques. These methods encompass classical approaches like GAT as well as SOTA methods tailor-made for addressing the heterophily problem, such as GBKSage and HOG. Additionally, we also compare our approach with the spectral method GPRGNN and local spectral methods FAGCN and ACM-GCN in solving the heterophily problem due to their widely validated effectiveness and code accessibility.

As for ACM+ (using layer normalization) and ACM++ (using residual and adjacency matrix), we believe that these generic tricks fail to truly showcase the model's inherent capabilities, while incorporating adjacency matrices suffers from potential data leakage. For conciseness, we select ACM-GCN as our baseline approach. 
The reason for choosing GloGNN instead of GloGNN++ is the same.

Similar to NCGCN, UDGNN also proposes a framework based on separating nodes. However, the lack of open-source code and the complexity of this method hinder its reproducibility. Moreover, direct comparison with reported results is challenging due to different splitting strategies employed across datasets in the original paper.

\subsection{\textbf{Parameter Tuning}}\label{app:parameter}
Considering the scale of datasets, we use Tesla A100 40/80GB for model training and parameter tuning. To optimize hyperparameters according to validation accuracy for all methods, we use the open-source toolkit NNI~\cite{nni2021} along with its default TPE~\cite{bergstra2011algorithms} algorithm.
\subsubsection{\textbf{Criteria for Hyperparameter Tunning}}
Based on our dataset selection criteria, many of the selected datasets have not been fully optimized by these baselines. To ensure a fair comparison, we provide detailed information on hyperparameter tuning as follows.

The GBKSage and HOG methods exhibit significant computational inefficiency, requiring more than 100 times longer computation time compared to other approaches. Consequently, we only utilized their default parameters in our experiments. 
As for CPGNN, reproducing its code under our dataset splits and random seeds demands substantial effort. Regrettably, we were unable to find an interface for hyperparameter tuning, such as adjusting the learning rate.

For other methods, we fine-tune the common hyperparameters within the search space, including learning rate, weight decay, and dropout (as presented in Table~\ref{tab: common-para}). Since these methods often involve multiple specific hyperparameters and their reported optimal settings did not yield satisfactory results in our experimental setup, we adopt the default configurations provided in their respective source codes or papers. 
Please refer to Appendix~\ref{app:hyper-baseline} for further details.

\subsubsection{\textbf{Common Parameters}}\label{app:comm-para}
Table~\ref{tab: common-para} shows the searching space of three common parameters, including learning rate, weight decay and dropout rates.
\begin{table}[!ht]
    \renewcommand\arraystretch{1.2}
    \centering
    \caption{Searching space for three common parameters of all methods.}
    \resizebox{0.9\linewidth}{!}{
    \begin{tabular}{ll} 
    \hline\hline
    \textbf{Parameter}        & \multicolumn{1}{c}{\textbf{Searching space}}                                            \\ 
    \hline
    learning rate    & \multicolumn{1}{c}{\{1e-3, 5e-3,  1e-2, 5e-2, 0.1\}}  \\
    weight decay     & \multicolumn{1}{c}{\{0, 5e-5, 1e-4, 5e-4, 1e-3\}}                                             \\
    dropout          & \multicolumn{1}{c}{[0, 0.9] with 0.1 interval}                             \\
    \hline\hline
    \end{tabular}}
    \label{tab: common-para}
\end{table}

\subsubsection{{\textbf{Specific Parameters for Competitors}}}\label{app:hyper-baseline}
All models follow the default parameter settings on their public codes.
For GPRGNN, we set the order of the graph filter (frequency-domain filters often use finite orders) to 10 and the dropout rate before the propagation of GPR weights to 0.5, and let the coefficients $\gamma_k$ be initialized by PPR and set the $\alpha$ to 0.5. 
For FAGCN, the residual coefficient $\epsilon$ of the initial feature is set to 0.3. Following the original setting, the number of layers is 4 for homophilous graphs and 2 for heterophilous ones.
For WRGAT, the relations of the multi-relational computation graph are set to 10. 
For GBKSage, the hyper-parameter $\lambda$ to balance two losses is set to 30. 
For SNGNN, we set the top $k=10$, threshold $thr=0.5$, $\alpha=0.5$ and the number of layers is 1.
For HOG, the threshold $r=0.8$ and $r_2=0.1$.
For CPGNN, we use the best variant CPGNN-Cheby-1 according to their reported accuracies.

\subsubsection{\textbf{Specific Parameters for NCGCN}}
Our NCGCN has only three specific parameters, as shown in Table~\ref{tab: specific-para}.
\textit{Hop} is chosen from $\{1, 2\}$ because we consider the \textit{NC} of a node as a local metric, and different graphs have different applicability to locality assumption. 
\textit{Add self-loop} refers to whether or not a self-loop is included in the GCN-like aggregation, i.e., \textit{Yes} means adopting 
$\operatorname{norm}(\boldsymbol{A})=(\boldsymbol{D}+\boldsymbol{I})^{-1/2}(\boldsymbol{A}+\boldsymbol{I})(\boldsymbol{D}+\boldsymbol{I})^{-1/2} $ 
and \textit{No} means adopting 
$\operatorname{norm}(\boldsymbol{A})=\boldsymbol{D}^{-1/2}\boldsymbol{A}\boldsymbol{D}^{-1/2} $. 
% The search space of specific hyperparameters is shown in Table~\ref{tab: specific-para}.

\begin{table}[!ht]
    \renewcommand\arraystretch{1.2}
    \centering
    \caption{Searching Space for Specific Parameters of NCGCN.}
    \resizebox{0.9\linewidth}{!}{
    \begin{tabular}{ll} 
    \hline\hline
    \textbf{Parameter}           & \multicolumn{1}{c}{\textbf{Searching space}}                                      \\ 
    \hline
    Hop $k$                & \multicolumn{1}{c}{\{1, 2\}}                                                              \\
    Threshold $T$           & \multicolumn{1}{c}{\{0.3, 0.4, 0.5, 0.6, 0.7\}}\\   
    Add self-loop            & \multicolumn{1}{c}{\{Yes, No\}}                                                            \\    
    \hline\hline
    \end{tabular}}
    \label{tab: specific-para}
\end{table}

% \newpage

\subsubsection{{\textbf{Optimal Parameters}}}
Our NCGCN includes both common and specific parameters, as shown in Table~\ref{tab: common-para} and Table~\ref{tab: specific-para}, respectively. The optimal parameters, obtained through parameter tuning on the validation set, are reported in Table~\ref{tab:NCGCN} and Table~\ref{tab:NCSAGE}.
In most datasets, parameter tuning mostly concludes with a different dropout rate between the low- and high-\textit{NC} group, indicating that different model complexity is required for different groups. 
The finding reflects the capability of NCGCN/NCSAGE to adaptively adjust model complexity for different node groups, aligning with observation 2 in Section~\ref{sec:NC}.

\begin{table*}[ht]
\renewcommand\arraystretch{1.2}
\centering
\caption{{Optimal parameters for NCGCN}}
\label{tab:NCGCN}
\resizebox{\textwidth}{!}{
\begin{tabular}{c|ccccc|ccccc} 
\hline\hline
                & \textbf{Pubmed} & \textbf{Photo} & \textbf{Computers} & \begin{tabular}[c]{@{}c@{}}\textbf{Coauthor} \\\textbf{CS}\end{tabular} & \begin{tabular}[c]{@{}c@{}}\textbf{Coauthor} \\\textbf{Physics}\end{tabular} & \textbf{Cora Full} & \textbf{Actor} & \textbf{Penn94} & \begin{tabular}[c]{@{}c@{}}\textbf{Chameleon}\\\textbf{Filtered}\end{tabular} & \begin{tabular}[c]{@{}c@{}}\textbf{Squirrel}\\\textbf{Filtered}\end{tabular}  \\ 
\hline
learning rate   & 0.1             & 1e-2& 5e-2& 1e-3& 5e-2& 1e-3& 0.1& 1e-3& 0.1& 0.1\\
weight decay    & 1e-4            & 5e-5& 5e-5               & 0& 1e-3& 1e-3               & 5e-3& 5e-5            & 5e-3& 5e-3\\
dropout (low)   & 0.4             & 0& 0.2& 0.9& 0.5                                                                          & 0
& 0.2& 0.7& 0.6& 0.3\\
dropout (high)  & 0.3             & 0.9& 0.6& 0.9& 0.9& 0.9& 0.6& 0& 0.2                                                                             & 0.5\\
hops $k$      & 1               & 1              & 2& 1                                                                       & 1                                                                            & 2                  & 1              & 1               & 2& 2                                                                             \\
add self-loop   & 1               & 1& 0& 1& 1& 1                  & 1              & 1               & 1                                                                             & 1                                                                             \\
threshold $T$ & 0.5             & 0.3                & 0.7& 0.7& 0.6                                                                          & 0.6& 0.7& 0.7             & 0.6& 0.3\\
\hline\hline
\end{tabular}}
\end{table*}

\begin{table*}[ht]
\renewcommand\arraystretch{1.2}
\centering
\caption{{Optimal parameters for NCSAGE}}
\label{tab:NCSAGE}
\resizebox{\textwidth}{!}{
\begin{tabular}{c|ccccc|ccccc} 
\hline\hline
                & \textbf{Pubmed} & \textbf{Photo} & \textbf{Computers} & \begin{tabular}[c]{@{}c@{}}\textbf{Coauthor} \\\textbf{CS}\end{tabular} & \begin{tabular}[c]{@{}c@{}}\textbf{Coauthor} \\\textbf{Physics}\end{tabular} & \textbf{Cora Full} & \textbf{Actor} & \textbf{Penn94} & \begin{tabular}[c]{@{}c@{}}\textbf{Chameleon}\\\textbf{Filtered}\end{tabular} & \begin{tabular}[c]{@{}c@{}}\textbf{Squirrel}\\\textbf{Filtered}\end{tabular}  \\ 
\hline
learning rate   & 0.1             & 5e-2& 1e-3& 1e-2                                                                     & 0.1                                                                          & 1e-3& 5e-2& 5e-3& 0.1& 0.1                                                                           \\
weight decay    & 1e-4            & 5e-5           & 5e-5& 0                                                                    & 5e-3                                                                         & 5e-5               & 5e-3 & 5e-5            & 5e-3& 5e-3\\
dropout (low)   & 0.2& 0.9& 0& 0.7                                                                     & 0.4& 0.5& 0.9& 0.7& 0.9& 0.8\\
dropout (high)  & 0.1& 0.7& 0.5& 0.8& 0.9& 0.3& 0.2& 0.1& 0.3& 0.5\\
hops $k$      & 1               & 2              & 2& 1& 1                                                                            & 2                  & 1              & 1               & 2& 2\\
add self-loop   & 1               & 0& 1& 1&0 & 1                  & 1              & 1               & 1                                                                             & 1\\
threshold $T$ & 0.4& 0.3& 0.3& 0.6& 0.4& 0.4& 0.6& 0.6             & 0.7& 0.4                                                                           \\
\hline\hline
\end{tabular}}
\end{table*}

\begin{table*}
    \renewcommand\arraystretch{1.4}
    \centering
    \caption{Results on real-world benchmark datasets reported in the original papers: 
    Mean Test Accuracy (\%) $\pm$ Standard Deviation. Boldface letters mark the best results while underlined letters indicate the second best. ``*'' means we adopt the best accuracy among a series of models they proposed.
    ``-'' means that the original paper did not report results under 60\%/20\%/20\% split.}
    \label{tab: public-leaderboard}
    \resizebox{\linewidth}{!}{
    \begin{tabular}{c|ccccccc} 
    \hline\hline
            & \textbf{Pubmed}              & \textbf{Photo}           & \textbf{Computers}    & \textbf{Coauthor CS}  & \textbf{Coauthor Physics} & \textbf{Cora Full}    & \textbf{Actor}        \\
    \hline       
    Geom-GCN~\cite{pei2019geom}            & 90.72                    & -                     & -                     & -                     & -                         & -                     & 31.96                 \\
    GBKSage~\cite{du2022gbk}               & 89.11 ± 0.23             & -                     & -                     & -                     & -                         & -                     & 38.97 ± 0.97          \\
    GloGNN*\cite{li2022glognn}             & 89.24 ± 0.39             & -                     & -                     & -                     & -                         & -                     & 37.70 ± 1.40          \\
    GCNII*~\cite{chen2020gcnii}            & 90.30                    & -                     & -                     & -                     & -                         & -                     & 41.54 ± 0.99          \\
    GPRGNN~\cite{chien2020adaptive}        & 89.18 ± 0.15             & -                     & -                     & -                     & -                         & -                     & 39.30 ± 0.27          \\
    CPGNN~\cite{zhu2021cpgnn}              & 82.44 ± 0.58             & -                     & -                     & -                     & -                         & -                     & -                     \\
    ACM*~\cite{luan2022acm}                & \underline{91.44 ± 0.59} & -                     & -                     & -                     & -                         & -                     & 41.86 ± 1.48          \\
    APPNP~\cite{APPNP}                     & 85.02 ± 0.09             & -                     & -                     & -                     & -                         & -                     & 38.86 ± 0.24        \\
    Snowball*~\cite{luan2019snowball}      & 89.04 ± 0.49             & -                     & -                     & -                     & -                         & -                     & 36.00 ± 1.36        \\
    CoLinkDist~\cite{luo2021colink}        & 89.58                    & 94.36                 & 89.42                 & \underline{95.80}     & 97.05                     & \underline{70.32}     & -                     \\
    BernNet~\cite{he2021bernnet}           & 88.48 ± 0.41             & 93.63 ± 0.35          & 87.64 ± 0.44          & -                     & -                         & -                     & 41.79 ± 1.01          \\
    3ference~\cite{3ference}               & 88.90                    & 95.05                 & 90.74                 & 95.99                 & \underline{97.22}         & -                     & -                     \\
    Exphormer~\cite{shirzad2023exphormer}  & -                        & \underline{95.35}     & \textbf{91.47}        & 94.93                 & 96.89                     & -                     & -                     \\
    FavardGNN~\cite{Fav}  & 90.90 ± 0.27   & -                        & -                     & -                     & -                     & -                         & \underline{43.05 ± 0.53}        \\
    \hline
    NCGCN/NCSAGE      & \textbf{91.64 ± 0.53} & \textbf{95.93 ± 0.36} & \underline{90.81 ± 0.46} & \textbf{96.64 ± 0.29} & \textbf{98.69 ± 0.26}    & \textbf{73.42 ± 0.58} & \textbf{43.89 ± 1.33} \\
    \hline\hline
    \end{tabular}}
\end{table*}

\begin{table*}[!htb]
    \centering
    \renewcommand\arraystretch{1.4}
    \caption{Average running time per epoch (ms) and average running time in a run(s).}
    \label{tab: effeciency}
    \resizebox{\textwidth}{!}{
    \begin{tabular}{c|cccc|cccc|cccccccc} 
        \hline\hline
        \textbf{Time on Pubmed} & MLP  & GCN  & SAGE &GAT  & H2GCN & GPRGNN & FAGCN & ACM-GCN & CPGNN  & WRGAT  & GBKSage & HOG     & GloGNN & SNGNN  & CAGNN & NCGCN  \\ 
        \hline
        time per epoch/ms       & 2.78 & 5.97 & 31.90 & 13.32 & 9.36  & 13.87  & 14.04 & 10.56   & 102.43 & 294.13 & 8109.29 & 6127.86 & 25.85  & 5.01   &14.14 
& 59.93  \\
        time per run/s          & 0.70 & 1.28 & 7.41 & 3.01  & 2.73  & 3.13   & 3.31  & 2.62    & 51.21  & 62.91  & 2286.01 & 1009.87 & 2.58   & 1.18  & 7.07 
& 11.41  \\
        \hline\hline
    \end{tabular}}
\end{table*}

\subsection{Performance Comparison on Public Leaderboard}\label{app: public-leaderboard}
% We compare our results with those provided in the original papers of baselines, as shown in Table~\ref{tab:full}. These models, including spectral methods, spatial methods and transformer-based methods, are currently the best methods under splitting of 60\%/20\%/20\% for training/validation/testing. Remarkably, our method NCGCN outperforms all of them on \texttt{Pubmed}, \texttt{Photo}, \texttt{Coauthor CS}, \texttt{Coauthor Physics} and \texttt{Actor}. 
% For \texttt{Cora Full}, we give up the first place on this benchmark due to too few comparison objects.

We compare our results with those reported in several relevant papers, as shown in Table~\ref{tab: public-leaderboard}. The compared works involve spectral methods, spatial methods and transformer-based methods, all of which represent currently the best methods under the 60\%/20\%/20\% data split setting for training/validation/testing. As can be seen, our methods achieve the best results in 6 out of the 7 datasets.

\subsection{More Analysis on Efficiency Comparison}\label{app:eff}

Table~\ref{tab: effeciency} reports the relevant time consumption data, corresponding to Fig.~\ref{fig: time}.
Here we explain the reason for OOM in Table~\ref{tab:main}.
\begin{itemize}
    \item \textbf{HOG:} utilizes a homophily degree matrix calculated based on the similarity between each pair of nodes. This \textbf{dense matrix} has a space complexity of $\mathcal{O}(n^2)$. Additionally, it incorporates three optimization objectives into the loss function.
    \item \textbf{GBKSage:} employs Kernel Selection Gate, which is a MLP used for training and predicting whether the neighboring node belongs to the same class as the central node. As a result, it introduces additional space complexity of $\mathcal{O}(|E|d)$, where $d$ represents the dimension of features or intermediate embeddings, leading to OOM issues in relatively dense graphs such as \texttt{Penn94}. 
    \item \textbf{WRGAT:} use multiple computation graphs (default: 10 graphs in their paper) to add structure information, which is equivalent to training 10 GATs in parallel. Additionally, experiments or discussions on large graphs are absent in their paper.
    \item \textbf{SAGE:} For full-batch SAGE, its space complexity is $\mathcal{O}(n|\mathcal{N}^2| d)$, which highly depends on the number of nodes in the 2-hop neighborhood. Note that \texttt{Penn94} is denser than other benchmarks, with an average degree of 66. Its size of 1-hop neighbors is about $66n$, but 2-hop is prohibitively high at $66^2n$. However, in the second layer of our framework, the average degree of two separated graphs is about 32, leading to a complexity of $2*32^2n$. Although NCSAGE doubles its parameters by $d^2$ due to two learners, it's neglectable compared to $n$. Finally, NCSAGE effectively reduces the parameter size by about $50\%$.
\end{itemize}

\end{document}